\documentclass[journal,comsoc]{IEEEtran}

\usepackage[T1]{fontenc}%

\ifCLASSINFOpdf
\else
\fi

\usepackage{amsmath}
\usepackage[cmintegrals]{newtxmath}

\usepackage[font=footnotesize,labelfont={sc,bf},textfont=footnotesize]{caption}

\usepackage{multirow}
\usepackage{adjustbox}
\usepackage{tabu}
\usepackage{tabularx}
\usepackage{xcolor}
\usepackage{soul}
\usepackage[linesnumbered, ruled]{algorithm2e}
\usepackage{algpseudocode}%
\usepackage{bm}
\usepackage{array}
\usepackage{enumitem}
\usepackage{verbatim}
\usepackage{algorithmicx}
\usepackage{comment}
\usepackage{lipsum}
\usepackage{stringstrings}
\usepackage{subfigure}
\usepackage{hyperref}

\def\eg{\textit{e.g.}}
\def\ie{\textit{i.e.}}

\def\etc{\textit{etc}}
\def\aka{\textit{a.k.a.}}

\def\bM{\mathbf{M}}
\def\bF{\mathbf{F}}
\def\bw{\mathbf{w}}
\def\dong{\textcolor{black}}
\def\xinyu{\textcolor{black}}
\def\xinyutwo{\textcolor{black}}
\def\check{\textcolor{black}}

\makeatletter
  \newcommand\figcaption{\def\@captype{figure}\caption}
  \newcommand\tabcaption{\def\@captype{table}\caption}
\makeatother

\begin{document}
\title{Part-Guided Attention Learning for Vehicle \\
Instance Retrieval}

\author{Xinyu~Zhang$^{*}$,~%
        Rufeng~Zhang$^{*}$,~%
        Jiewei~Cao,~%
        Dong~Gong,~%
        Minyu~You$^{\#}$,
        ~Chunhua Shen~%
\thanks{Manuscript received April 06, 2020; revised September 24, 2020.
}
\thanks{X. Zhang, R. Zhang and M. You are with Department of Control Science and Engineering, Tongji University, Shanghai 201804, China (e-mail: zhangxinyu@tongji.edu.cn; cxrfzhang@tongji.edu.cn; myyou@tongji.edu.cn). M. You is also with Shanghai Institute of Intelligent Science \& Technology, Tongji University, Shanghai 201804, China.}
\thanks{J. Cao, D. Gong and C. Shen are with The University of Adelaide, Adelaide, SA 5005, Australia (e-mail: jonbakerfish@gmail.com; edgong01@gmail.com; chunhua.shen@adelaide.edu.au).
JC, DG, CS and
 their employer received no financial support for the research, authorship, and/or publication of this article.
}
\thanks{${^*}$Part of this work was done when X. Zhang was visiting The University of Adelaide. First two authors contributed to this work equally.}
\thanks{${^\#}$Correspondence should be addressed to M. You.}%
}

\markboth{
}%
{
}
\maketitle

\begin{abstract}
Vehicle instance retrieval (IR) often requires one to recognize the fine-grained visual differences between vehicles.
Besides the holistic appearance of vehicles which is easily affected by the viewpoint variation and distortion, vehicle parts also provide crucial cues to differentiate near-identical vehicles.
Motivated by these observations, we introduce a \textit{Part-Guided Attention Network} (PGAN) to pinpoint the prominent part regions and effectively combine the global and local information for discriminative feature learning.
PGAN first detects the locations of different part components and salient regions regardless of the vehicle identity, which serves as the \textit{bottom-up attention} to narrow down the possible searching regions.
To estimate the importance of detected parts, we propose a \textit{Part Attention Module} (PAM) to adaptively locate the most discriminative regions with high-attention weights and suppress the distraction of irrelevant parts with relatively low weights.
The PAM is guided by the identification loss and therefore provides \textit{top-down attention} that enables attention to be calculated at the level of car parts and other salient regions.
Finally, we aggregate the global
appearance and local features together to improve the feature performance further.
The PGAN combines part-guided bottom-up and top-down attention, global and local visual features in an end-to-end framework.
Extensive experiments demonstrate that the proposed method achieves new state-of-the-art vehicle IR performance on four large-scale benchmark datasets.\footnote{Code is available at: \href{https://git.io/PGAN-Vehicle}{ https://git.io/PGAN-Vehicle}.}
\end{abstract}


\IEEEpeerreviewmaketitle

\begin{figure}[t!]
\centering
\includegraphics[width=0.95\linewidth]{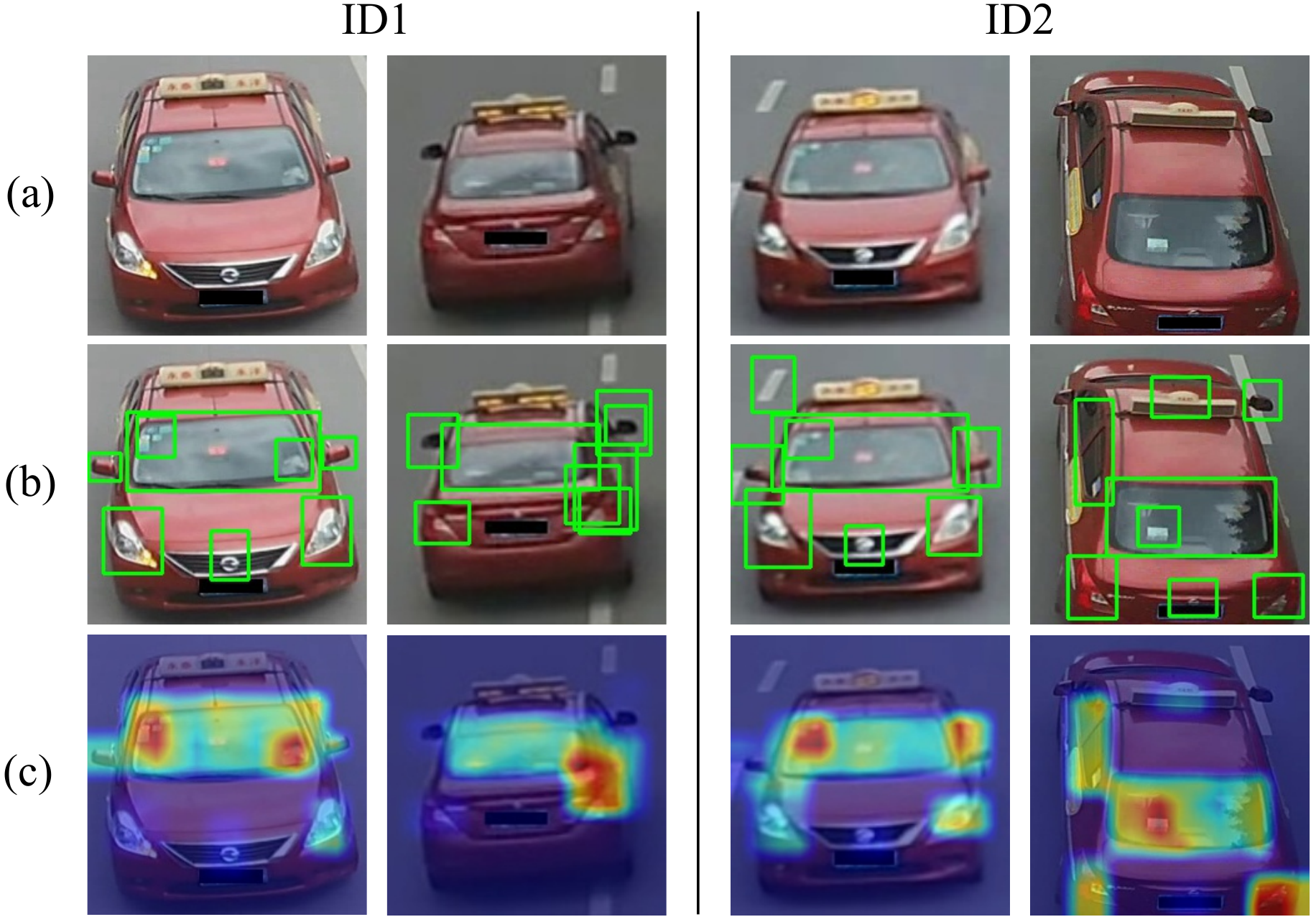}
\caption{%
Illustration of the part-guided attention.
(a) The rear and front views of two different vehicles with the same car model.
(b) The detected candidate part regions from the part extraction module.
(c) The heatmaps of part features from the part attention module. The prominent part regions like annual signs are highlighted, while the wrong candidates and insignificant parts like background and back mirror are suppressed.
}
\label{fig:figure1}
\end{figure}

\section{Introduction}
\IEEEPARstart{V}{ehicle} instance retrieval (IR) aims to verify whether or not two vehicle images captured by different cameras belong to the same identity.
Vehicle IR is also known as vehicle re-identification. With the growth of road traffic, it plays an increasingly important role in urban systems and intelligent transportation~\xinyu{\cite{arth07,feris2012large,compcars,liu2016large,wang2017orientation,vehicle1M,cityflow,veriwild,zheng2018fast,zheng2020vehiclenet}}.

Different levels of granularity of visual attention are required under various IR scenarios.
In the case of comparing vehicles of different car models, we can easily distinguish their identities by examining the overall appearances, such as car types and headlights~\cite{compcars}.
However, most production vehicles %
can exhibit
near-identical appearances since they may be mass-produced by the same manufacturer.
When two vehicles with the same car model are presented, more fine-grained details (\eg, annual service signs, customize paintings, and personal decorations) are required for comparison, as shown in Figure~\ref{fig:figure1} (a) that ID$1$ looks similar like ID$2$ since they are from the same car mode.
Therefore, the key challenge of vehicle IR lies in how to recognize the subtle differences between vehicles and locate the prominent parts that characterize their identities.

Most existing works focus on learning global appearance features with various vehicle attributes, including model type~\cite{vehicle1M,liu2016deep,wei2018coarse}, license plate~\cite{liu2016deep}, spatial-temporal information~\cite{liu2017provid,shen2017learning}, orientation~\cite{wang2017orientation,zhou2017cross,zhou2018aware,lou2019embedding}, \etc.
The main disadvantage of global features is the lack of capability to capture more fine-grained visual differences, which is crucial in vehicle IR.
Despite the help of auxiliary attributes, the supervision is still weak.
For instance, the license plates are usually not available for
privacy protection, while the two extremely similar vehicles from the same model type can not be distinguished (as shown in Figure~\ref{fig:figure1}).
Also, they are easily degraded by the viewpoint variation, distortion, occlusion, motion blur and illumination, especially in the unconstrained real-world environment.
Therefore, it is important to explore more robust and environment-invariant information to represent specific vehicles.
Recent many works tend to explore subtle variances from car parts~\cite{he2019part,zhu2019vehicle,zhao2019structural}
to learn the local information.
However, these methods mainly focus on the localization of the spatial part regions without
considering how these regions are subject to attention with different degree.

To address above problems, we propose a novel \textit{part-guided attention network} (PGAN) to improve the performance effectively by focusing on the most prominent part regions, which is implemented by integrating the bottom-up attention and the top-down attention systematically.
Specially, we first utilize a bottom-up attention module to extract the related vehicle part regions, which is called the \textit{part extraction module} in our work.
With the established object detectors~\cite{compcars,zhao2019structural} that are pre-trained on the vehicle attributes, we consider the extracted part regions from the part extraction module as candidates,
which is beneficial for narrowing down the searching area for network learning.
Importantly, this bottom-up attention can effectively take advantage of the context correlation among pixels in the same part via assigning same values for all pixels in a specific part region, which is superior to grid attention that gives no consideration on the pixel relationships.
Besides, we call these candidate regions as \textit{coarse part regions} since the quality of the detection may be not accurate with the pre-trained part extraction module and some part regions with less information may be included in these candidates.

To extract more effective local information, we apply a top-down attention process to select the most prominent part regions as well as assign appropriate importance scores to them after obtaining the above candidate part regions.
Here, we introduce a \textit{part attention module} (PAM), which is guided by the identification loss to allocate the importance for each coarse part region.
PAM adaptively locates the discriminative regions with high-attention weights and suppresses the distraction of irrelevant parts with relatively low weights, as shown in Figure~\ref{fig:figure1} (c).
It is also beneficial for filling out the wrongly detected part regions by giving a weight near to zero (as shown in the rear view of ID2 in Figure~\ref{fig:figure1}).
In detail, PAM can assign a special attention weight for each corresponding part region, \ie, all pixels in this region share the same weight, reflecting the importance of the selected part regions by considering all pixels in a part region as a whole.
Therefore, PAM is more efficient than grid attention or evenly decomposed part attention~\cite{chen2019partition,chen2019multi,zhu2019vehicle,liu2018ram}, since PAM
is able to provide more fine-grained attention which is conducted only on the selected part regions by taking the context information among pixels into consideration instead of all spatial pixels. We call these selected-weighted part regions as \textit{fine part regions}.

\begin{figure}[t!]
\centering
\includegraphics[trim =0mm 0mm 0mm 0mm, clip, width=0.9\linewidth]{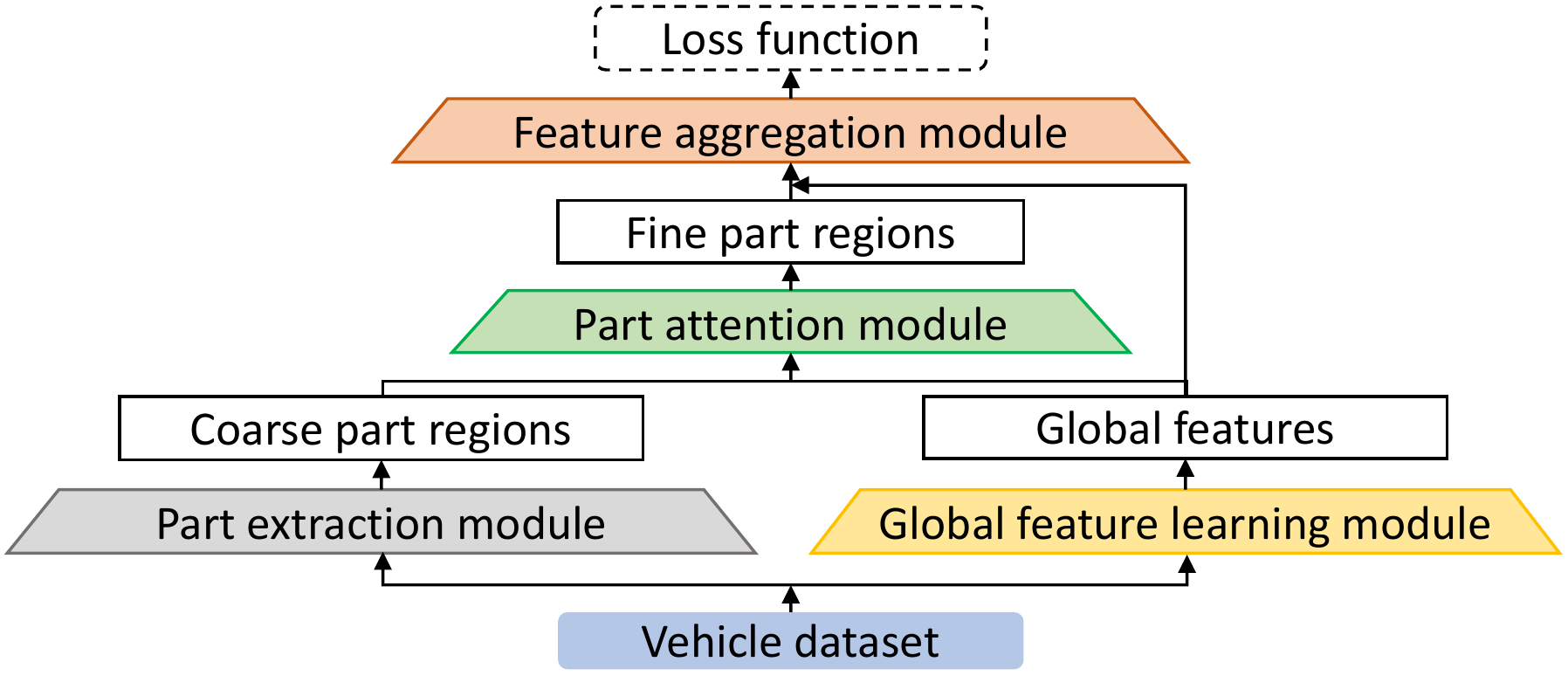}
\caption{
The flow of the
overall
training process of our framework. The trapezoid represents the modules involved in our framework, while the solid rectangle denotes the outputs of the modules.}
\label{fig:figure2}
\end{figure}

With the combination of bottom-up and top-down attention, our attention mechanism can provide more prominent part regions for improving the feature representation.
Finally, we aggregate the vehicle's holistic appearance and part characteristics with a \textit{feature aggregation module} to improve the performance further.
Figure~\ref{fig:figure2} shows the whole training process.%
To summarize, our main contributions are as follows:
\setlist{nolistsep}
\begin{itemize}[fullwidth, itemindent=1em]
\item We design a novel Part-Guided Attention Network (PGAN), which effectively combines part-guided bottom-up and top-down attention together to capture both local and global information.
\item We propose to extract Top-$D$ part regions without part alignment to maintain more prominent yet less available parts effectively in the part extraction module.
\item We propose a part attention module (PAM) to evaluate the relative importance of the selected Top-$D$ part regions, which further focuses more on prominent parts and reducing the distraction of wrongly detected or irrelevant parts.
\item Extensive experiments on four challenging benchmark datasets demonstrate that our proposed method achieves new state-of-the-art vehicle IR performance.
\end{itemize}
\setlist{nolistsep}

\begin{figure*}[t]
\centering
\includegraphics[width=0.95\linewidth]{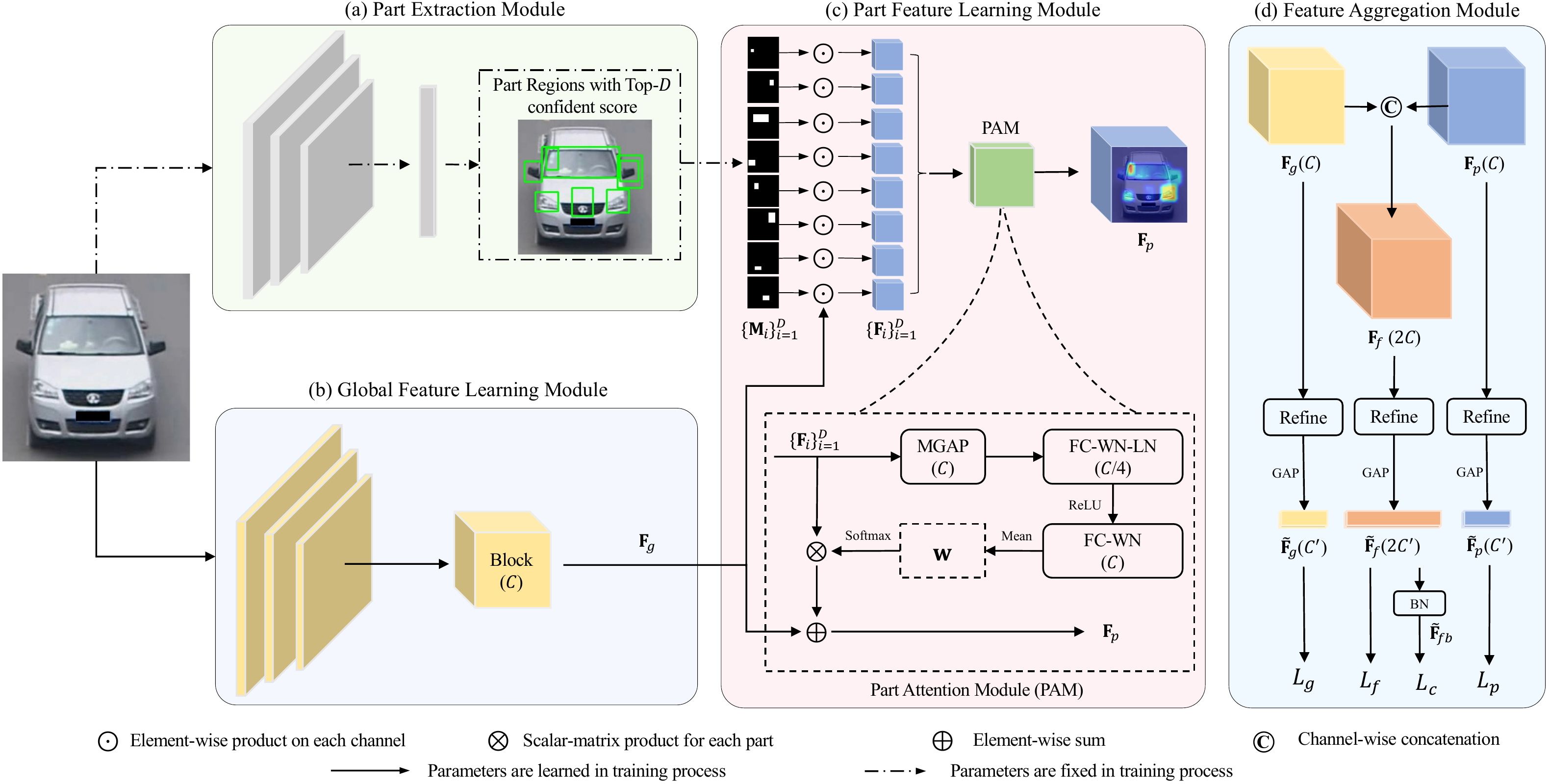}  %
\caption{Part-Guided Attention Network (PGAN) pipeline.
The model consists of four modules: Part Extraction Module, Global Feature Learning Module, Part Feature Learning Module and Feature Aggregation Module.
The input vehicle image is first processed to obtain the global feature $\mathbf{F}_{g}$ and the part masks $\{\mathbf{M}_{i}\}_{i=1}^D$ of Top-$D$ candidate parts.
The part mask features $\{\mathbf{F}_i\}_{i=1}^D$ is then obtained via Eq.~\eqref{eq:mask},
after which $\{\mathbf{F}_i\}_{i=1}^D$ is fed into a Part Attention Module (PAM) to obtain the part-guided feature $\mathbf{F}_{p}$.
PAM is a
compact
network, learning a soft attention weight $\mathbf{w} \in \mathbb{R}^{D}$, which is
composed of a mask-guided average pooling (MGAP) layer and some linear and non-liner layers.
Subsequently, the fusion feature $\mathbf{F}_f$ is obtained by concatenating $\mathbf{F}_g$ and $\mathbf{F}_p$. After the refinement and global average pooling (GAP) operation, $\widetilde{\mathbf{F}}_g$, $\widetilde{\mathbf{F}}_p$ and $\widetilde{\mathbf{F}}_{f}$ are all used for
\xinyu{the optimization of triplet loss functions $L_f$, $L_g$ and $L_p$, respectively}.
\xinyu{Besides, as \cite{luo2019bag}, $\widetilde{\mathbf{F}}_{f}$ is followed by a BN layer and the normalized feature $\widetilde{\mathbf{F}}_{fb}$ is used for optimizing the softmax cross-entropy loss $L_c$.}
Here, FC, WN, LN and BN represent fully-connected layer, weight normalization, layer normalization and batch normalization respectively. Mean denotes a channel-wise mean operation. $C$ and $C'$ are channel dimension before and after refine operation.}
\label{fig:framework}
\end{figure*}

\section{Related Work}
\subsection{Global Feature-based Methods}
\textbf{Feature Representation} Vehicle IR aims at learning discriminative feature representation to deal with significant appearance changes for different vehicles. Public large-scale datasets~\cite{compcars,liu2016large,vehicle1M,liu2016deep,veriwild,pkuvehicleid,vric} are widely collected with annotated labels and abundant attributes under unrestricted conditions. These datasets face huge challenges on occlusion, illumination, low resolution and various views. One way to deal with these datasets uses deep features~\cite{liu2016large,wang2017orientation,liu2016deep,tang2017multi,vric} instead of hand-crafted features to describe vehicle images. To learn more robust features, some methods~\cite{vehicle1M,liu2016deep,wei2018coarse,liu2017provid,shen2017learning,liu2018progressive} try to explore details of vehicles using additional attributes, such as model type, color, spatial-temporal information,  \etc.
Moreover, works of \cite{zhou2017cross,lou2019embedding} propose to use synthetic multi-view vehicle images from a generative adversarial network (GAN)~\cite{goodfellow2014generative} to alleviate cross-view influences among vehicles. In~\cite{wang2017orientation,zhou2018aware} authors also implement view-invariant inferences effectively by learning a viewpoint-aware representation.
Although great progress has been obtained by these methods, there is a huge drop when encountering invisible variances of different vehicles as well as large diversities in the same vehicle identity.\\
\textbf{Metric Learning} To alleviate the above limitation, deep metric learning methods~\cite{yan2017exploiting,sanakoyeu2019divide,kumar2019vehicle,yuan2017hard} use powerful distance metric expression to pull vehicle images in the same identity closer while pushing dissimilar vehicle images further away.
The core idea of these methods is to utilize the matching relationship between image pairs or triplets as much as possible,
\xinyutwo{which are widely used in IR works~\cite{batchhardtriplet,lin2019group,luo2019bag}}.
Whereas, sampling strategies in deep metric learning lead to suboptimal results and also lack of abilities to recognize more meaningful unobtrusive details.
It is thus limited by the complex differences of the vehicle appearances.

\subsection{Part Feature-based Methods}
\xinyutwo{Similar as \cite{lin2017learning,sun2018beyond,zhang2019progressive,wang2018learning,li2019amur} focusing on object patches in other IR works, a series of part-based learning methods explicitly exploit the discriminative information from multi-part locations of vehicles.
}
\xinyu{\cite{zhao2019structural} provides an attribute detector while \cite{cui2017vehicle} provides a marker detector.}
\cite{zhu2019vehicle,chen2019partition,chen2019multi,liu2018ram},\xinyu{~\cite{liu2019group}} take great efforts on separating feature maps into multiple even partitions to extract specific features of respective regions.
However, it is difficult for vehicles to directly apply this naive partitions since the vehicle appearances change a lot.
In other words, almost all pedestrian images have relative regular appearances from top to bottom (representing head to feet), while vehicle appearances suffer from multiple views without unique commons.
For example, the bottom partition of the front vehicle is wheels, while that is the vehicle back of the rear one in ID2 in Figure~\ref{fig:figure1}.
Although \cite{zhu2019vehicle} utilizes discriminative features from quadruple directions for each vehicle,
it still suffers from the misalignment problem due to inaccurate grid partitions.

Another line of part-based methods \xinyu{\cite{wang2017orientation}},~\cite{khorramshahi2019dual,kanaci2019multi,khorramshahi2019attention} bring informative key-points to put more attention on effective localized features.
In particular, although \cite{khorramshahi2019dual} attempts to detect and use keypoints,
it defines a heuristic rule to choose keypoints for every input image.
Actually, \cite{khorramshahi2019dual} extracts a sub-group of keypoint features based on the vehicle orientation, in which the choice of the keypoint groups is manually pre-defined.
\cite{wang2017orientation} applies an aggregation module on the local features based on orientation. However,
keypoints in each orientation are treated equally and the detail information is easy to be ignored.
In contrast, we adaptively learn a soft-attention for each detected local part feature, conditional on the input image.
The soft-attention coefficients, measuring the importance of a local feature for the target task, \xinyu{are learnt by} using the sole target identification loss.
In other word, we do not rely on
extra information while \cite{khorramshahi2019dual} uses orientation as extra supervision.

Besides, ~\cite{he2019part,zhao2019structural} denote to design part-fused networks using ROI features of each part on vehicles from a pre-trained detection model to extract discriminative features.
\xinyu{
However, there is no importance selection on the candidate part regions in \cite{he2019part}, which considers all part regions equally.
Instead, our PGAN can select the most prominent part regions, \eg, annual service
signs and hungs, which are subtle yet important to distinguish different vehicles.
In addition, we apply a single tailor-designed supervision for the soft-weighted part features together.
Compared with \cite{he2019part} applying separate supervision to each part feature, our PGAN can provide more accurate supervision with the aggregated feature.
}
\xinyu{Although \cite{teng2018scan,teng2020multi,guo2019two} also utilize attention in the feature maps, the attention mechanism is applied on each pixel in the feature map.
Our part attention module focuses on the pixel sets, \ie, detected part regions on the feature maps.
Thus, the context correlation in a same part can be integrally
considered.}
In this way, we can not only consider all part features together as a whole but also pay more attention to the prominent part regions as well as alleviate the influence of irrelative ones.

\section{Methodology}
We firstly define each vehicle image as $x$ and the unique corresponding identity label as $y$. Given a training set $X^t$=$\{(x^t_n, y^t_n)\}_{n=1}^{N^t}$, the main goal of the vehicle IR is to learn a feature embedding function $\phi (x^{t};~\mathbf{\theta})$ for measuring the vehicle similarity under certain metrics, where $\theta$ denotes the parameters of $\phi(\cdot)$.
It is important to learn a $\phi$ with good generalization on unseen testing images since there is no overlap identities in training and testing dataset.
During testing, given a query vehicle image $x^q$, we can find vehicles with the same identity from a gallery set $X^g$ = $\lbrace(x^g_n, y^g_n)\rbrace^{N_{g}}_{n=1}$ by comparing the similarity between $\phi (x^q; \mathbf{\theta})$ and each $\phi (x^g_n; \mathbf{\theta}),~\forall x_n^g$.

In this section, we present the proposed Part-Guided Attention Network (PGAN) in detail. The overall framework is illustrated in Figure~\ref{fig:framework}, which consists of four main components: \textit{Part Extraction Module}, \textit{Global Feature Learning Module}, \textit{Part Feature Learning Module} and \textit{Feature Aggregation Module}.
We first generate the part masks of vehicles in the part extraction module, which are then applied on the global feature map to obtain the mask-guided part feature.
After that, we learn the attention scores of different parts to enhance the part feature via increasing the weights of discriminative parts as well as decreasing that of less informative parts.
Subsequently, the three refined features, \ie, global, part, and fusion features are all used for model optimization.

\subsection{Global Feature Learning Module}
For a vehicle image $x$, before obtaining the part features, we first extract a global feature map $\mathbf{F}_g\in \mathbb{R}^{H\times W \times C}$ with a standard convolutional neural network, as shown in Figure \ref{fig:framework} (a).
Most previous methods~\cite{batchhardtriplet,luo2019bag} directly feed $\mathbf{F}_g$ into a global average pooling (GAP) layer to obtain the embedding feature that mainly considers the global information, which is studied as a \textit{baseline} model in our experiments.

However, due to the lost of the spacial information after GAP, it is difficult to distinguish two near-identical vehicles, as illustrated in ID1 and ID2 in Figure~\ref{fig:figure1}.
Therefore, it is crucial to maintain the spatial structure of feature maps, which helps describe the subtle visual differences.
We thus directly apply $\mathbf{F}_{g}$ as one of the inputs for the following part learning process and the final optimization, and we explore a novel method to focus on the effective part regions following.

\subsection{Part Extraction Module}
We first extract the part regions using a pre-trained SSD detector specially trained on vehicle attributes \cite{zhao2019structural}.%
Here, we only consider $16$ of all $21$ vehicle attributes as shown in Table~\ref{table:attributes}.
The reason is that the remaining attributes are vehicle styles, \ie, ``car'', ``trunk'', ``tricycle'', ``train'' and ``bus'', representing the whole vehicle image
which can be recognized as the global information in our paper.
\dong{Once detected, we only use the confidence scores to select part regions and ignore the label information of each part.
It is reasonable since not all attributes are available in each vehicle due to the multi-view variation, so that
it is hard to
decide a universal rule for reliable part alignments (\ie, selecting same part regions for all vehicles).
}

\begin{table}[t]
\caption{Name and abbreviation of vehicle attributes used in our paper.}
\label{table:attributes}
\centering
\setlength{\tabcolsep}{0.5mm}{
\xinyu{\begin{tabular}{l|c|l|c} %
\hline
Name & Abbreviation & Name & Abbreviation \\
\hline
\hline
annual service signs & anusigns & back mirror & backmirror \\
car light & carlight & carrier & carrier\\
car topwindow & cartopwindow & entry license & entrylicense \\
hanging & hungs & lay ornament & layon\\
light cover & lightcover & logo & logo \\
newer sign & newersign & tissue box & tissuebox \\
plate & plate & safe belt & safebelt \\
wheel & wheel & wind-shield glass & windglass\\
\hline
\end{tabular}}
}
\end{table}

Instead of naively %
selecting relevant part regions by setting a threshold on the confidence scores, we select the most confident top-$D$ proposals as the candidate vehicle parts. The main reasons are twofold:
1) some crucial yet less confident bounding boxes, like annual service signs, play a crucial role in distinguishing different vehicle images;
2) part number is fixed, which is easy to learn the attention model in the following stage.
\dong{Note that we want to ensure a high recall rate to avoid missing relevant parts. The irrelevant parts are filtered out from the subsequent top-down attention learning.
}

We use the index $i\in\{1,2,...,D\}$ to indicate each of the selected top-$D$ part regions. The spatial area covered by each part is denoted as $A_i$. For each candidate part region $i$, we obtain a binary mask matrix $\bM_i\in\{0,1\}^{H\times W}$ by assigning $1$ to the elements inside the part region $A_i$ and $0$ to the rest, denoted as:
\begin{equation}
\centering
\begin{aligned}
{\bM}_{i}(\mathrm{pix}) &= \left\{\begin{matrix}
1, ~~~\text{if}~~\mathrm{pix} \in A_{i}\\
0, ~~~\text{if}~~\mathrm{pix} \notin A_{i}
\end{matrix}\right., \forall i,  \\
\end{aligned}
\end{equation}
where $\mathrm{pix}$ indicates a pixel location of $\bM_i$.
Note that the size of each $\bM_i$ is the same as a single channel of $\mathbf{F}_g$.
It means that if the parameters of the neural network or the sizes of input images change, the corresponding part locations on $\bM_{i}$ will be changed accordingly and the spacial area $A_i$ is also changed.
Although $\bM$ can be scaled based on the input of multi-scale images, we resize all images to the same resolution for simplification and thus the size of $\bM$ can be regularized to ${H\times W}$.
Besides, during processing, we force all $A_i$ in the range of $H\times W$ to ensure all part regions are located in the range of image areas (\ie, the size of ${H\times W}$).
After obtaining global feature $\mathbf{F}_g$ and part masks $\{\bM_i\}_{i=1}^D$, we project the part masks on the feature map $\mathbf{F}_g$ to generate a set of mask-based part feature representations $\{\mathbf{F}_i\}_{i=1}^D$, which will be taken as the input of the following part feature attention module.
For each part region $i$, we can obtain $\mathbf{F}_i$ via the following formula:
\begin{equation}
\centering
\mathbf{F}_{i} = \bM_{i}\odot \mathbf{F}_{g}, ~~\forall i\in \{1,2,...,D\},
\label{eq:mask}
\end{equation}
where $\odot$ denotes the element-wise product operation on each channel of $\mathbf{F_g}$. $\mathbf{F}_{i}$ is the mask-based part feature map of the $i$-th part region. Note that all $\mathbf{F}_i\in\mathbb{R}^{H\times W\times C}$. In each $\bF_i$, only the elements in the regions of $i$-th part are activated.
The illustration is shown in Figure~\ref{fig:framework} \xinyu{(c)}.
\dong{We learn an attention module on the part regions in the following section.
Unlike the traditional grid attention method that processes a set of uniform grids, our attention model can focus on the prominent parts by only activating the selected parts. The irrelevant parts can thus be ignored directly.
Besides, the context correlation in a same part can be integrally considered,
alleviating missing of essential features.
Moreover,
this part extraction process can be considered as a bottom-up attention mechanism~\cite{anderson2018bottom} with a set of candidate images regions proposed.
}

\subsection{Part Feature Learning Module}
Part feature learning module is to produce a weight map across the mask-based part feature maps $\{\bF_i\}$.
In this way, the network can focus on specific part regions.
\xinyu{Recent methods~\cite{he2019part,wang2019vehicle} highlight all part regions equally
and thus
ignores the importance discrepancy among different part regions.}
\dong{
Besides,
some detected parts might not be informative for some specific cases, such as wrongly detected background or windshield without useful information, which tends to result in degraded results.}
\dong{To tackle the above problems, we propose a \textit{part attention module} (PAM) to adaptively learn the importance of each part so as to take more attention to the most discriminating
regions and suppress those with less information. %
Consequently, PAM can be considered as a part-based top-down attention mechanism,
since this attention signal is supervised by the specific identification task
to predict
an importance distribution over candidate image regions.}

\begin{figure}[t!]
\centering
\includegraphics[trim =-2mm 0mm 0mm 0mm, clip, width=0.875\linewidth]{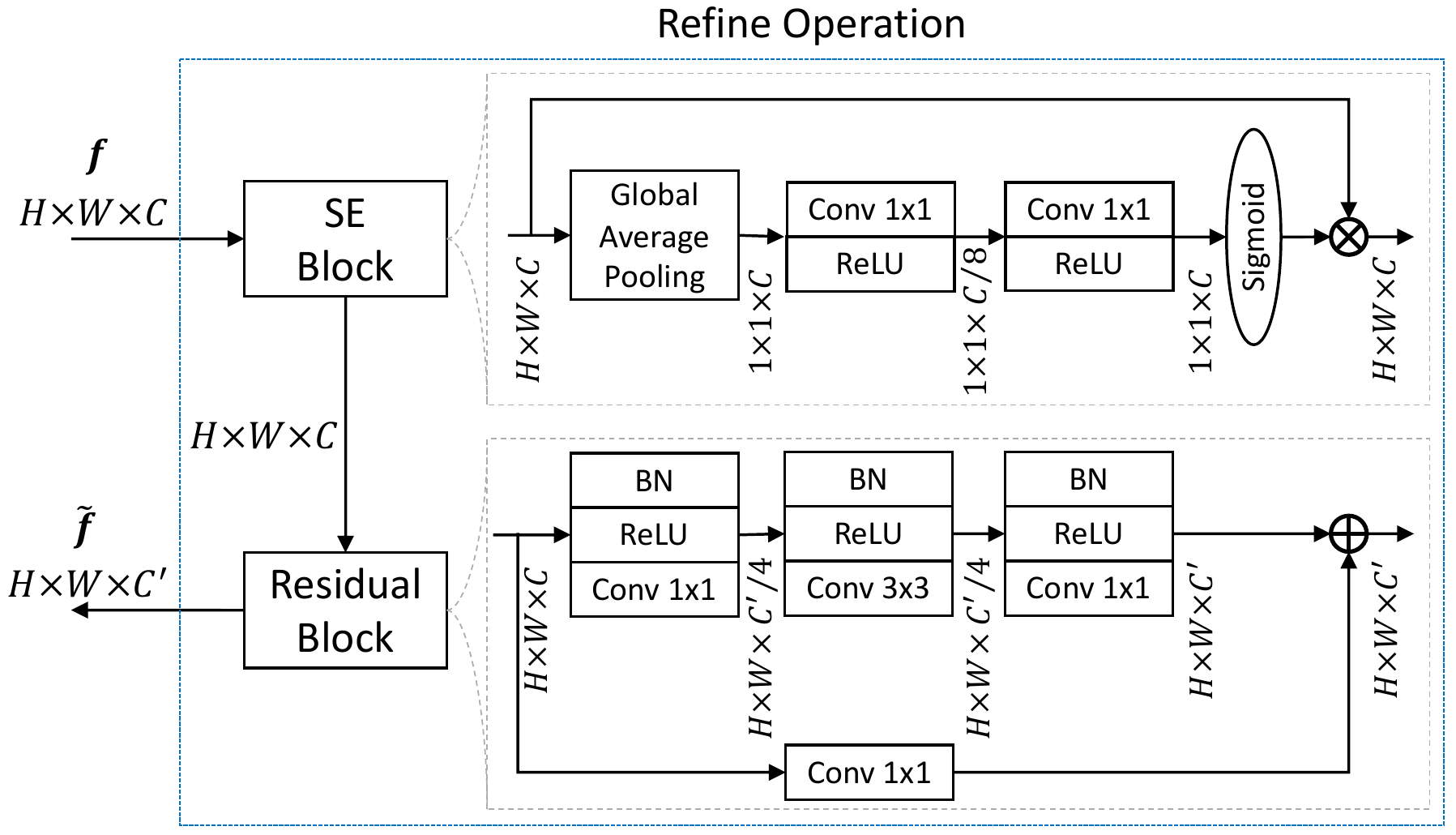}
\caption{
\xinyu{The module structure in the refine operation. $C$ and $C'$ are the size of channel of feature maps before and after the refine operation.}
}
\label{fig:refinemodule}
\end{figure}

\noindent \textbf{Part attention module (PAM)}~
Our PAM is designed to obtain a part-guided feature representation $\mathbf{F}_{p}\in \mathbb{R}^{H\times W \times C}$ relying on a top-down attention mechanism on candidate part regions.
From PAM, we can obtain a soft weight vector $\mathbf{w}\in \mathbb{R}^{D}$
to indicate the importance of each part region,
thus the part-guided feature representation $\bF_p$ can be obtained as:
\begin{equation}
    \mathbf{F}_p = \sum_{i=1}^{D}w_{i}\mathbf{F}_{i} + \mathbf{F}_{g},\\
\end{equation}
where $w_i\in[0,1]$ denotes the $i$-th element of the soft weight $\mathbf{w}$, which represents a learned weight of $i$-th part feature $\mathbf{F}_{i}$ obtained via Eq.~\eqref{eq:PAM}. $\bw$ is normalized with sum as 1 so that the relative importance between different parts is obvious.
Here, $\mathbf{F}_{g}$ is added to augment the capability of part regions.

We learn a compact model to predict the attention weights $\bw$ for measuring the different importance of each selected part, as shown in Figure \ref{fig:framework} (c). Specifically, we first use a mask-guided global average pooling operation (MGAP) on each $\mathbf{F}_i$ and then learn a mapping function with a softmax layer to obtain $\bw$. Each element $w_i$ can be predicted by:
\begin{equation}
\centering
\begin{aligned}
w_i = \frac{{\rm exp}(~\psi({\rm mgap}(\bF_i, \bM_i), \theta_\psi ) ~)}{\sum_{j=1}^{D}{\rm exp}(~\psi({\rm mgap}(\bF_j, \bM_j), \theta_\psi ) ~)}, \\
\end{aligned}
\label{eq:PAM}
\end{equation}
where $\psi\left( \cdot \right)$ denotes a learnable function that is able to highlight the most important part regions with high values (as shown in Figure \ref{fig:framework} (c)).
$\theta_\psi$ is the parameter of mapping function $\psi(\cdot)$, and $\mathrm{mgap}(\cdot)$ denotes
MGAP operation discussed in the following.

\par
Before feeding $\bF_i$ into $\psi$, we average each channel of $\bF_i$ as a scalar via the $\mathrm{mgap}(\cdot)$ operator. Note that, in each $\bF_i$, only the elements in the part region $i$ are activated and most of the elements in $\bF_i$ are zero. Instead of performing the standard global average pooling (GAP), we restrict the average pooling in the areas indicated by the mask $\bM_i$ via the MGAP operator. In detail, for each channel of $\bF_i$, after summing the nonzero elements, the MGAP operator devides the sum value with the number of elements (\ie $||\bM_i||_1 < H\times W $), instead of the number of total elements
(\ie ~$H\times W$) in the GAP.

\subsection{Feature Aggregation Module}
Since global and part-based features provide complementary information, we concatenate the global feature $\mathbf{F}_{g}$ and part-guided feature $\mathbf{F}_{p}$ together, which is then denoted as fusion feature $\mathbf{F}_{f}\in \mathbb{R}^{H\times W \times 2C}$.
Furthermore,
we adopt a \textit{Refine} operation on $\mathbf{F}_{f}$ to reduce the dimension of feature representation to speed up the training process.
The \textit{Refine} operation is composed of a SE Block~\cite{hu2018squeeze} and a Residual Block~\cite{resnet50},
\xinyu{which is illustrated in Figure~\ref{fig:refinemodule}.}
\xinyu{After a global average pooling (GAP) layer, the refined fusion feature $\widetilde{\mathbf{F}}_{f}\in \mathbb{R}^{{2C}'}$, $\widetilde{\mathbf{F}}_{g}\in \mathbb{R}^{{C}'}$ and $\widetilde{\mathbf{F}}_{p}\in \mathbb{R}^{{C}'}$ are obtained for the whole model optimization.}
\xinyu{Here, $C'$ is the size of channel of feature maps after the refine operation, while $C$ is that before the refine operation.}
\xinyu{Note that following~\cite{luo2019bag}, an additional batch normalization(BN) layer is adopted on $\widetilde{\mathbf{F}}_{f}$.
It is proved to be beneficial for optimizing the softmax cross-entropy loss~\cite{luo2019bag}.
Here, we denote the feature after the BN layer as $\widetilde{\mathbf{F}}_{fb}$.
}

\subsection{Model Training}
\xinyu{In the training process, we adopt softmax cross-entropy loss and triplet loss~\cite{batchhardtriplet} as a joint optimization.
In specific, we apply triplet loss on $\widetilde{\mathbf{F}}_{f}$ and softmax cross-entropy loss on $\widetilde{\mathbf{F}}_{fb}$,
denoted as $L_{f}$ and $L_{c}$.
In order to make full use of the global and part information separately, we also optimize the refined global feature $\widetilde{\mathbf{F}}_{g}$ and part-guided feature $\widetilde{\mathbf{F}}_{p}$ with triplet loss, which are denoted as $L_{g}$ and $L_{p}$, respectively.
}
Overall, the total loss function can be formulated as:
\begin{equation}
\begin{aligned}
L = \lambda L_c + L_{tri} = \lambda L_c + L_f + L_g + L_p,\\
\end{aligned}
\label{eq:loss}
\end{equation}
where $\lambda$ is the loss weight to trade off the influence of two types of loss functions, \ie, softmax cross-entropy loss $L_c$ and triplet loss $L_{tri}$.
Experiments show that joint optimization could improve the ability of feature representation.

\xinyu{For evaluation, we use the normalized fusion feature $\widetilde{\mathbf{F}}_{fb}$ as the final
feature representation in our work.}

\begin{table}[t]
\caption{\xinyu{Comparison on the different optimization methods of PGAN on VeRi-776.
$\widetilde{\mathbf{F}}_{fb}$ is used as the feature representation.
For fairness, the feature dimension is fixed to 512 for all methods including the baseline model.
We omit the loss weight in Eq.~\ref{eq:loss} for clarity and set $\lambda$ to 2 here.}
}
\label{table:optimization-analysis}
\centering
\setlength{\tabcolsep}{1.5mm}{
\xinyu{\begin{tabular}{l|c|c|c|c}
	\hline %
\multicolumn{1}{l|}{Method} & \multicolumn{1}{c|}{Optimization} & \multicolumn{1}{c|}{mAP} & \multicolumn{1}{c|}{Top-1} & \multicolumn{1}{c}{Top-5} \\
\hline
\hline
Baseline & $L_{c} + L_{f}$ & 75.7 & 95.2 & 98.2\\
\hline
\multirow{4}{*}{Ours} & $L_{c} + L_{f}$ & 77.7 & 95.9 & \textbf{98.5} \\
 & $L_{c} + L_{f} + L_{g}$ & 78.0 & 95.1 & 97.7 \\
 & $L_{c} + L_{f} + L_{p}$ & 78.5 & 95.8 & 98.3\\
 & $L_{c} + L_{f} + L_{g} + L_{p}$ (PGAN) & \textbf{79.3} & \textbf{96.5} & 98.3 \\ 	\hline
\end{tabular}}
}
\end{table}

\section{Experiments}\label{experiment}
\subsection{Datasets and Evaluation Metrics} \label{sec:dataset}
We evaluate our PGAN method on four public large-scale Vehicle IR (\aka, re-identification) benchmark datasets.

\textit{VeRi-776}~\cite{liu2016deep} is a challenging benchmark in vehicle IR task that contains about $50,000$ images of $776$ vehicle identities across $20$ cameras.
Each vehicle is from $2$-$18$ cameras with various viewpoints, illuminations and occlusions. All datasets are split into a training set with $37,778$ images of $576$ vehicles and a testing set with $11,579$ images with $200$ vehicles.

\textit{VehicleID}~\cite{pkuvehicleid} is a widely-used vehicle IR dataset which contains vehicle images captured in the daytime by multiple cameras. There are total of $221,763$ images with $26,267$ vehicles, where each vehicle has either front or rear view. The training set contains $110,178$ images of $13,134$ vehicles while the testing set comprises $111,585$ images of $13,133$ vehicles.
\xinyu{The evaluation protocol of the large test subset VehicleID is randomly selecting one image from each vehicle to generate a gallery set (2400 images) while the remaining images are used as query set. The random selection process was repeated for 10 times and the mean result is used as the final performance.}

\textit{VRIC}~\cite{vric} is a realistic vehicle IR benchmark with unconstrained variations of images in resolution, motion blur, illumination, occlusion, and multiple viewpoints. It contains $60,430$ images of $5,622$ vehicle identities captured from 60 different traffic cameras during both daytime and nighttime. The training set has $54,808$ images of $2,811$ vehicles, while the rest is used for testing with $5,622$ images of another $2,811$ vehicle IDs.

\textit{VERI-Wild}~\cite{veriwild} is recently released with $416,314$ vehicle images of $40,671$ IDs captured by $174$ cameras. The training set consists of $30,671$ IDs with $277,797$ images.
The small test subset consists of $3,000$ IDs with $41,816$ images while the medium and large subset consist of $5,000$ and $10,000$ IDs with $69,389$ and $138,517$ images respectively.

\textit{Evaluation metrics.} To measure the performance for vehicle IR, we utilize the Cumulated Matching Characteristics (CMC) and the mean Average Precision (mAP) as evaluation criterions. The CMC calculates the cumulative percentage of correct matches appearing before the top-$K$ candidates. We report Top-$1$ and Top-$5$ scores to represent the CMC criterion. Given a query image, Average Precision (AP) is the area under the Precision-Recall curve while mAP is the mean value of AP across all query images. The mAP criterion reflects both precision and recall, which provides a more convincing evaluation on IR task.

\subsection{Implementation Details}
\textit{Part extraction.}
\xinyutwo{
we directly conduct the inference process to extract part regions using the pretrained detector~\cite{zhao2019structural}.
There is no re-train or finetune process in our method since the attribute annotations in the four datasets, \ie, VeRi-776, VehicleID, VRIC and VERI-Wild, are not available.
In the training process of the detector in~\cite{zhao2019structural}, which is based on the SSD model~\cite{ssd}, the VOC21\_S dataset~\cite{zhao2019structural} is used as the training data.
The VOC21\_S dataset is captured during both daytime and nightime by multiple real-world cameras in several cities, so that this dataset shares similar scenarios with the four datasets we used.
Since these datasets are collected by different cameras in not exactly the same environment, there is a domain gap issue to some extent.
During the inference, the NMS threshold is set to 0.45 in all experiments.
For each image, we extract Top-$D$ part regions according to confident scores, where $D=8$ without specification.
}

\textit{Vehicle IR model.} We adopt ResNet50~\cite{resnet50} without the last classification layer as the backbone model in the global feature learning module, which is pre-trained on ImageNet~\cite{imagenet} initially.
The model modification follows~\cite{luo2019bag},
\xinyu{\ie, \xinyutwo{removing} the last downsample operation and adding a BN layer before softmax cross-entropy loss}.

All images are resized to $224\times 224$.
The data augmentations, \ie, random horizontal flipping and random erasing~\cite{randomerasing} with a probability of $0.5$, are used as in \cite{luo2019bag}.
We use Adam optimizer~\cite{adam}
with a momentum of $0.9$ and a weight decay $5\times 10^{-4}$. For all experiments without other specification, we set the batch size to $64$ with $16$ IDs randomly selected. The learning rate starts from $1.75\times 10^{-4}$ and is multiplied by $0.5$ every $20$ epochs. The total number of epochs is $130$.

\begin{table}[t!]
\caption{Performance comparison on different attention methods, \ie, grid attention, PGAN without Part Attention Module (PAM) and our PGAN on VeRi-776.}
\label{table:component-analysis}
\centering
\setlength{\tabcolsep}{2.0mm}{
\begin{tabular}{l|c|c|c|c}
	\hline %
\multicolumn{1}{l|}{Method} & \multicolumn{1}{c|}{Dimension} & \multicolumn{1}{c|}{mAP} & \multicolumn{1}{c|}{Top-1} & \multicolumn{1}{c}{Top-5} \\
\hline
\hline
Baseline & \multirow{4}{*}{256} & 75.3 & 95.3 & 98.2\\
Grid Attention &  & 76.1 & 95.3 & 97.7 \\
PGAN w/o PAM &  & 77.9 & 95.6 & \textbf{98.4} \\
PGAN &  & 78.6 & 95.4 & 98.0\\ \hline
Baseline & \multirow{4}{*}{512} & 75.7 & 95.2 & 98.2\\
Grid Attention &  & 77.0 & 95.8 & 98.0 \\
PGAN w/o PAM &  & 78.0 & 95.5 & 98.2\\
PGAN &  & \textbf{79.3} & \textbf{96.5} & 98.3\\  \hline
\end{tabular}}
\end{table}

\subsection{Ablation Study}
\subsubsection{Effectiveness of joint optimization}
We first design an ablation experiment analyzing the effectiveness of
\xinyu{joint optimization with different features and loss functions.}
\xinyu{For our method, we use the normalized feature $\widetilde{\mathbf{F}}_{fb}$ as the feature representation and fix the feature dimension of $\widetilde{\mathbf{F}}_{fb}$ to $512$, \ie, $C'=256$.}
For a fair comparison, we also set the feature dimension to $512$ in a baseline model.
As reported in Table~\ref{table:optimization-analysis}, we can observe that only using
\xinyu{optimization on the fusion feature, \ie, $L_c$ + $L_f$,}
can improve the performance by $2\%$ on mAP comparing with baseline model, which confirms that PAM can provide important part information that is better for model optimization.
After adding
\xinyu{$L_g$ and $L_p$}
separately, mAP can improve by about $1\%$.
\xinyu{It shows that combining with the additional optimizations on the global and part feature can provide more useful information for the model training.}
Furthermore, with the joint optimization with all these loss functions, the result improves to $79.3\%$ mAP, which outperforms the baseline model by $3.6\%$.

\begin{figure}[t!]
\begin{center}
\includegraphics[trim =0mm 0mm 0mm 0mm, clip, width=0.75\linewidth]{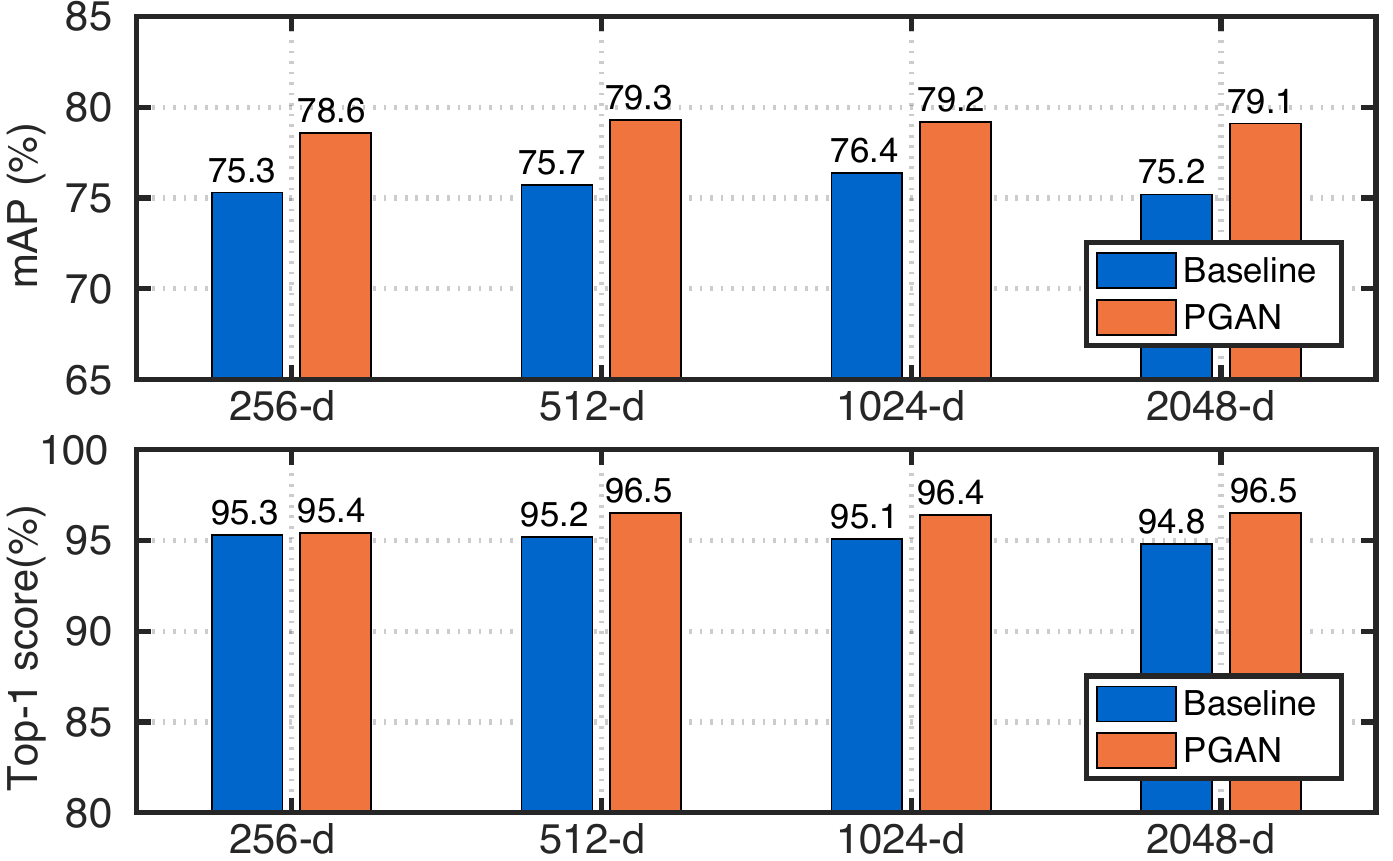}
\caption{Parameter analysis of the feature dimension on VeRi-776.}
\label{fig:dimension}
\end{center}
\end{figure}

\begin{figure}[t!]
\centering
\includegraphics[trim =0mm 0mm 0mm 0mm, clip, width=.75\linewidth]{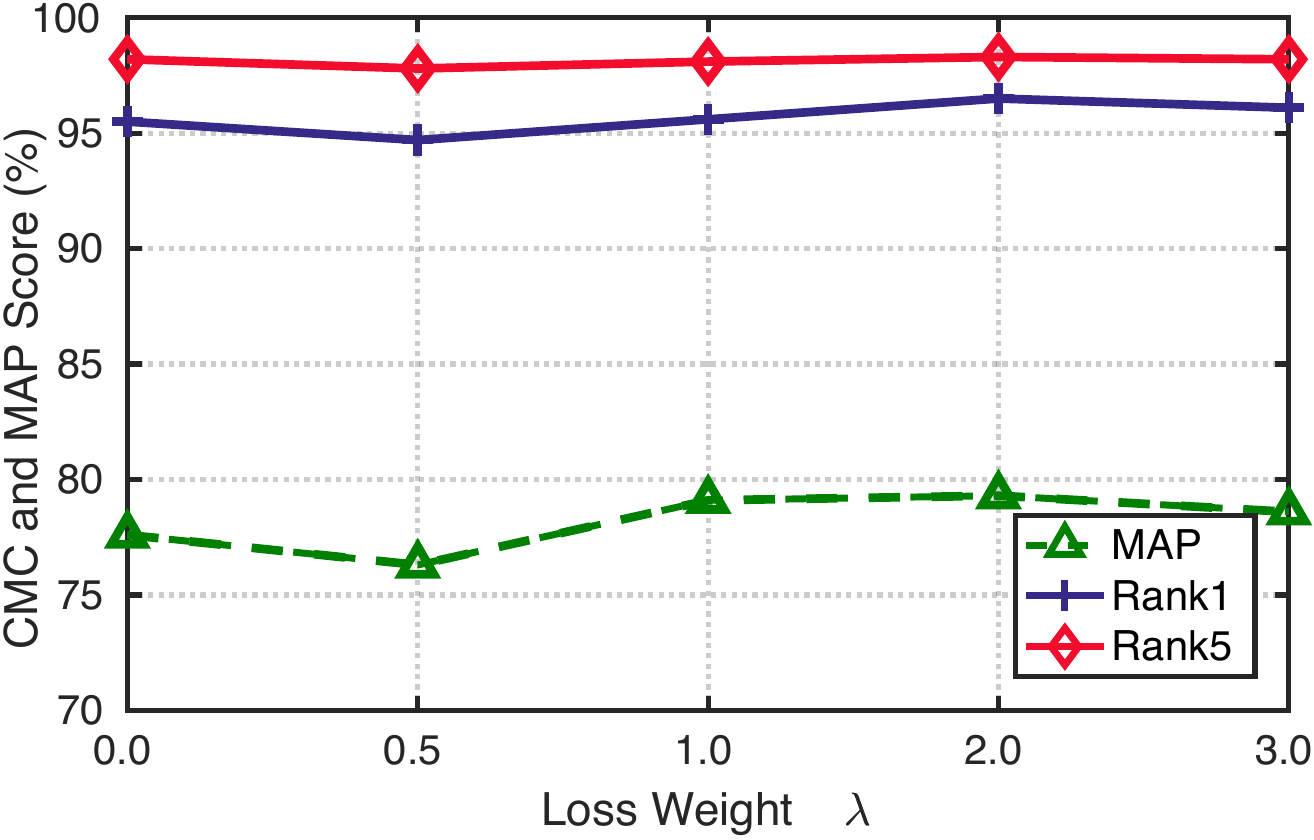}
\caption{Parameter analysis of the loss weight $\lambda$ on VeRi-776.}
\label{fig:lossweight}
\end{figure}

\subsubsection{Analysis of different attention method}
We first implement traditional grid attention by removing part extraction module, \ie, PAM is directly used on each grid of $\mathbf{F}_g\in \mathbb{R}^{H\times W\times C}$.
As shown in Table~\ref{table:component-analysis}, grid attention can only achieve $77.0\%$ mAP and $95.8\%$ Top-1 accuracy when the feature dimension is $512$, showing that part guidance is crucial for filtering invalid information like background.
Moreover, we also use the identical weight for each part region by removing PAM. It can be seen as a bottom-up attention with the part guidance from a detection model.
From Table~\ref{table:component-analysis}, we can find $0.7\%$ and $1.3\%$ mAP decrease when feature dimension is $512$ and $256$ without PAM.
It proves that PAM is beneficial for focusing on prominent parts as well as suppressing the impact of some wrongly detected or useless regions.
We exactly note that our PGAN w/o PAM is still better than grid attention by $1.0\%$ mAP, which also proves the important role of the part-guided bottom-up attention.

\subsubsection{Parameter analysis of the feature dimension}
We first analyze the effectiveness of different feature dimension.
The dimension ${2C}'$ of fusion feature $\widetilde{\mathbf{F}}_{fb}$ on VeRi-776 is used as the variable.
As shown in Figure~\ref{fig:dimension}, our PAM module has consistent improvement compared with the baseline model whatever the dimension is.
In particular, when $2C'=2048$, our PGAN outperforms baseline model by $3.9\%$ and $1.7\%$ in mAP and Top-$1$ respectively.
Besides, it is worth noting that our PGAN with low dimension still performs better than baseline with high dimension.
For example, our PGAN with $256$ dimension surpasses the baseline model with $512$ dimension by a large margin ($78.6\%$ \textit{vs.} $75.7\%$ mAP), which highly proves the effectiveness of our PGAN.

\subsubsection{Parameter analysis of the loss weight $\lambda$}
In Figure~\ref{fig:lossweight}, we conduct experiments to compare different values of the loss weight $\lambda$ in Eq.~\eqref{eq:loss}, which evaluates the trade off between softmax cross-entropy loss and triplet loss.
When $\lambda=0$, we only use triplet loss on $\widetilde{\mathbf{F}}_{f}$ as the optimization.
It is clear that when adding softmax cross-entropy loss on $\widetilde{\mathbf{F}}_{fb}$ into the model, our approach can obtain further improvement when the range of $\lambda$ is from $1.0$ to $2.0$.
However, too small $\lambda$ and too large $\lambda$ both lead to the bad influence for the model training.
We believe that there is a trade off between these two-type loss functions.
Too large $\lambda$ means the relation restriction among samples from the triplet loss does less effort for the model optimization, while too small $\lambda$ means less effectiveness of the global structure from the softmax cross-entropy loss.
Form Figure~\ref{fig:lossweight} we can see that the best result is obtained when $\lambda$ is set to $2$.
Without specification, we use $\lambda=2$ as the default loss weight in our paper.

\begin{table}[t!]
\caption{Performance comparison on different part number $D$ of PGAN on VeRi-776. $\widetilde{\mathbf{F}}_{fb}$ is used as the feature representation.}
\label{table:partnum-analysis}
\centering
\setlength{\tabcolsep}{1.5mm}{
\xinyu{
\begin{tabular}{l|c|c|c|c|c|c}
\hline %
\multicolumn{1}{l|}{\multirow{2}{*}{Part Number $D$}} & \multicolumn{3}{c|}{\multirow{1}{*}{Dimension 256}} &
\multicolumn{3}{c}{\multirow{1}{*}{Dimension 512}} \\
\cline{2-7}
 & \multicolumn{1}{c|}{mAP} & \multicolumn{1}{c|}{Top-1} & \multicolumn{1}{c|}{Top-5} & \multicolumn{1}{c|}{mAP} & \multicolumn{1}{c|}{Top-1} & \multicolumn{1}{c}{Top-5}\\
\hline
\hline
Baseline & 75.3 & 95.3 & 98.2 & 75.7 & 95.2 & 98.2\\
\hline
4 & 76.9 & 95.4 & 97.8 & 76.8 & 94.8 & 98.0 \\
6 & 78.5 & \textbf{95.8} & 98.0 & 78.7 & 96.2 & 98.0 \\  %
8 & \textbf{78.6} & 95.4 & 98.0 & \textbf{79.3} & \textbf{96.5} & \textbf{98.3}\\
10 & 77.6 & 95.5 & \textbf{98.3} & 79.1 & 95.9 & 98.2 \\  %
12 & 77.9 & 94.7 & 97.9 & 77.5 & 95.6 & 98.1 \\  \hline
\end{tabular}}
}
\end{table}

\begin{table}[t!]
\caption{Performance comparison on different part number $D$ of PGAN on VRIC and VERI-Wild (large subset). $\widetilde{\mathbf{F}}_{fb}$ is used as the feature representation. The feature dimension is fixed to 512.}
\label{table:partnum-analysis-vric-veriwild}
\centering
\setlength{\tabcolsep}{1.5mm}{
\xinyu{
\begin{tabular}{l|c|c|c|c|c|c}
\hline %
\multicolumn{1}{l|}{\multirow{2}{*}{Part Number $D$}} & \multicolumn{3}{c|}{\multirow{1}{*}{VRIC}} &
\multicolumn{3}{c}{\multirow{1}{*}{VERI-Wild (large)}} \\
\cline{2-7}
 & \multicolumn{1}{c|}{mAP} & \multicolumn{1}{c|}{Top-1} & \multicolumn{1}{c|}{Top-5} & \multicolumn{1}{c|}{mAP} & \multicolumn{1}{c|}{Top-1} & \multicolumn{1}{c}{Top-5}\\
\hline
\hline
Baseline & 83.5 & 76.1 & 93.0 & 69.4 & 88.1 & 95.4\\
\hline
4 & 84.1 & 76.8 & 93.0 & \textbf{70.8} & 89.5 & \textbf{95.9} \\
6 & 84.6 & 77.6 & 93.6 & 70.7 & 89.4 & 95.8 \\  %
8 & \textbf{84.8} & \textbf{78.0} & 93.2 & 70.6 & 89.2 & 95.7\\
10 & 84.1 & 76.8 & 93.2 & 70.7 & 89.6 & 95.8 \\  %
12 & 84.3 & 77.1 & \textbf{93.8} & 70.6 & \textbf{89.7} & 95.8 \\  \hline
\end{tabular}}
}
\end{table}

\begin{table*}[t!]
\caption{Comparisons with state-of-the-art IR methods on VeRi-776, VehicleID, VRIC and VERI-Wild. In each column, the first and second highest results are highlighted by {\bf{\color{red} red}} and {\color{blue} blue} respectively. The results of Siamese-CNN+Path-LSTM~\cite{shen2017learning} and OIFE~\cite{wang2017orientation} on VRIC is reported by MSVR~\cite{vric}.
$*$ denotes that VANet uses ResNet50 for VehicleID dataset while uses GoogLeNet for VeRi.}
\label{table:SOTA}
\begin{center}
\setlength{\tabcolsep}{0.7mm}{
\begin{tabular}{l|c|cc|ccc|ccc|c|c|c|c|c|c}
\hline
\multirow{3}{*}{Method} & \multicolumn{1}{c|}{\multirow{3}{*}{Backbone}} & \multicolumn{2}{c|}{\multirow{2}{*}{VeRi-776}} & \multicolumn{3}{c|}{\multirow{2}{*}{VehicleID}} & \multicolumn{3}{c|}{\multirow{2}{*}{VRIC}} & \multicolumn{6}{c}{VERI-Wild} \\
\cline{11-16}
& & & &   & &  &  &  & & \multicolumn{2}{c|}{Small} & \multicolumn{2}{c|}{Medium}  & \multicolumn{2}{c}{Large}\\
\cline{3-16}
& & \multicolumn{1}{c|}{mAP} & \multicolumn{1}{c|}{Top-1} & \multicolumn{1}{c|}{\xinyu{mAP}} & \multicolumn{1}{c|}{Top-1}   & \multicolumn{1}{c|}{Top-5} & \multicolumn{1}{c|}{\xinyu{mAP}} & \multicolumn{1}{c|}{Top-1} & \multicolumn{1}{c|}{Top-5} & mAP & Top-1 & mAP   & Top-1 & mAP   & Top-1\\
\hline
\hline
FACT+Plate-SNN+STR~\cite{liu2016deep} & \xinyu{GoogleNet} & \multicolumn{1}{c|}{27.8} &  \multicolumn{1}{c|}{61.4} &  \multicolumn{1}{c|}{-} & \multicolumn{1}{c|}{-} & \multicolumn{1}{c|}{-} & \multicolumn{1}{c|}{-} & \multicolumn{1}{c|}{-} & - & - & - & - & - & - & - \\ %
Siamese+Path-LSTM~\cite{shen2017learning} & \xinyu{ResNet50} & \multicolumn{1}{c|}{58.3} & \multicolumn{1}{c|}{83.5} & \multicolumn{1}{c|}{-} & \multicolumn{1}{c|}{-} & \multicolumn{1}{c|}{-} &  \multicolumn{1}{c|}{-} & \multicolumn{1}{c|}{30.6} & \multicolumn{1}{c|}{57.3} & - & - & - & - & - & -\\ %
OIFE~\cite{wang2017orientation} & \xinyu{GoogleNet} & \multicolumn{1}{c|}{51.4} & \multicolumn{1}{c|}{92.4} & \multicolumn{1}{c|}{-} & \multicolumn{1}{c|}{67.0} & \multicolumn{1}{c|}{82.9} & \multicolumn{1}{c|}{-} & \multicolumn{1}{c|}{24.6} & \multicolumn{1}{c|}{51.0} & - & - & - & - & - & -\\ %
PROVID~\cite{liu2017provid} & \xinyu{GoogleNet} & \multicolumn{1}{c|}{53.4} & \multicolumn{1}{c|}{81.6} & \multicolumn{1}{c|}{-} & \multicolumn{1}{c|}{-} & \multicolumn{1}{c|}{-} &  \multicolumn{1}{c|}{-} & \multicolumn{1}{c|}{-} & \multicolumn{1}{c|}{-} & - & - & - & - & - & -\\ %
VAMI~\cite{zhou2018aware} & \xinyu{Self-design} & \multicolumn{1}{c|}{50.1} & \multicolumn{1}{c|}{77.0} & \multicolumn{1}{c|}{-} & \multicolumn{1}{c|}{47.3} & \multicolumn{1}{c|}{70.3} & \multicolumn{1}{c|}{-} & \multicolumn{1}{c|}{-} &  - & - & - & - & - & - & -\\  %
MSVR~\cite{vric} & \xinyu{MobileNet}& \multicolumn{1}{c|}{49.3} & \multicolumn{1}{c|}{88.6} & \multicolumn{1}{c|}{-} & \multicolumn{1}{c|}{63.0} & \multicolumn{1}{c|}{73.1} &  \multicolumn{1}{c|}{-} & \multicolumn{1}{c|}{\color{blue}46.6} & \multicolumn{1}{c|}{\color{blue}65.6} & - & - & - & - & - & - \\ %
RNN-HA~\cite{wei2018coarse}& \xinyu{ResNet50} & \multicolumn{1}{c|}{56.8} & \multicolumn{1}{c|}{74.8} & \multicolumn{1}{c|}{-} & \multicolumn{1}{c|}{\bf {\color{red}81.1}} & \multicolumn{1}{c|}{\color{blue}87.4} & \multicolumn{1}{c|}{-} & \multicolumn{1}{c|}{-} & - & - & - & - & - & - & -\\ %
\xinyu{SCAN~\cite{teng2018scan}} & \xinyu{VGG16} & \multicolumn{1}{c|}{49.9} & \multicolumn{1}{c|}{82.2} & \multicolumn{1}{c|}{-} & \multicolumn{1}{c|}{65.4} & \multicolumn{1}{c|}{78.5} & \multicolumn{1}{c|}{-} & \multicolumn{1}{c|}{-} & - & - & - & - & - & - & -\\ %
RAM~\cite{liu2018ram} & \xinyu{VGGM} & \multicolumn{1}{c|}{61.5} & \multicolumn{1}{c|}{88.6} & \multicolumn{1}{c|}{-} & \multicolumn{1}{c|}{67.7} & \multicolumn{1}{c|}{84.5} & \multicolumn{1}{c|}{-} & \multicolumn{1}{c|}{-} & - & - & - & - & - & - & -\\ %
AAVER~\cite{khorramshahi2019dual} & \xinyu{ResNet50} & \multicolumn{1}{c|}{66.4} & \multicolumn{1}{c|}{90.2} & \multicolumn{1}{c|}{-} & \multicolumn{1}{c|}{63.5} & \multicolumn{1}{c|}{85.6} & \multicolumn{1}{c|}{-} & \multicolumn{1}{c|}{-} & - &  - & - & - & - & - & -\\ %
Part-Regular~\cite{he2019part} & \xinyu{ResNet50} & \multicolumn{1}{c|}{\color{blue}74.3} & \multicolumn{1}{c|}{\color{blue}94.3} &
\multicolumn{1}{c|}{-} & \multicolumn{1}{c|}{74.2} & \multicolumn{1}{c|}{86.4} & \multicolumn{1}{c|}{-} & \multicolumn{1}{c|}{-} & - & - & - & - & - & - & -\\ %
FDA-Net~\cite{veriwild} & \xinyu{Self-design} & \multicolumn{1}{c|}{55.5} & \multicolumn{1}{c|}{84.3} & \multicolumn{1}{c|}{61.8} & \multicolumn{1}{c|}{55.5} & \multicolumn{1}{c|}{74.7} & \multicolumn{1}{c|}{-} & \multicolumn{1}{c|}{-} & - & {\color{blue}35.1} & {\color{blue}64.0} & {\color{blue}29.8} & {\color{blue}57.8} & {\color{blue}22.8} & {\color{blue}49.4}\\ %
QD-DLF~\cite{zhu2019vehicle} & \xinyu{Self-design} & \multicolumn{1}{c|}{61.8} & \multicolumn{1}{c|}{88.5} & \multicolumn{1}{c|}{68.4} & \multicolumn{1}{c|}{64.1} & \multicolumn{1}{c|}{83.4} & \multicolumn{1}{c|}{-} & \multicolumn{1}{c|}{-} & \multicolumn{1}{c|}{-} & \multicolumn{1}{c|}{-} & \multicolumn{1}{c|}{-} & \multicolumn{1}{c|}{-} & - & \multicolumn{1}{c|}{-} & \multicolumn{1}{c}{-}\\ %
VANet~\cite{Chu_2019_ICCV}$^*$ & \xinyu{ResNet50} & \multicolumn{1}{c|}{66.3} & \multicolumn{1}{c|}{89.8} & \multicolumn{1}{c|}{-} & \multicolumn{1}{c|}{\color{blue}80.4} & \multicolumn{1}{c|}{\bf {\color{red}93.0}} & \multicolumn{1}{c|}{-} & \multicolumn{1}{c|}{-} & - &  - &- & - & - & - & -\\ %
\xinyu{TAMR~\cite{guo2019two}} & \xinyu{ResNet18} & \multicolumn{1}{c|}{-} & \multicolumn{1}{c|}{-} & \multicolumn{1}{c|}{61.0} & \multicolumn{1}{c|}{59.7} & \multicolumn{1}{c|}{73.9} & \multicolumn{1}{c|}{-} & \multicolumn{1}{c|}{-}
& - & - & - & - & - & - & -\\ %
\xinyu{GRF+GGL~\cite{liu2019group}} & \xinyu{VGGM} & \multicolumn{1}{c|}{61.7} & \multicolumn{1}{c|}{89.4} & \multicolumn{1}{c|}{-} & \multicolumn{1}{c|}{70.0} & \multicolumn{1}{c|}{87.1} & \multicolumn{1}{c|}{-} & \multicolumn{1}{c|}{-}
& - & - & - & - & - & - & -\\ %
\xinyu{MVAN~\cite{teng2020multi}} & \xinyu{ResNet50} & \multicolumn{1}{c|}{72.5} & \multicolumn{1}{c|}{92.6} & \multicolumn{1}{c|}{76.8} & \multicolumn{1}{c|}{72.6} & \multicolumn{1}{c|}{83.1} & \multicolumn{1}{c|}{-} & \multicolumn{1}{c|}{-}
& - & - & - & - & - & - & -\\ %
\hline
\xinyu{Baseline~\cite{luo2019bag}} & \xinyu{ResNet50} & \multicolumn{1}{c|}{75.7} & \multicolumn{1}{c|}{95.2} & \multicolumn{1}{c|}{83.5} & \multicolumn{1}{c|}{77.5} & \multicolumn{1}{c|}{91.0} &
\multicolumn{1}{c|}{83.5} & \multicolumn{1}{c|}{76.1} & \multicolumn{1}{c|}{93.0} &
\multicolumn{1}{c|}{82.6} & \multicolumn{1}{c|}{94.0} &
\multicolumn{1}{c|}{77.2} & \multicolumn{1}{c|}{91.7} &
\multicolumn{1}{c|}{69.4} & \multicolumn{1}{c}{88.1}\\
\hline
PGAN & \xinyu{ResNet50} & \multicolumn{1}{c|}{\bf{\color{red} 79.3}} &\multicolumn{1}{c|}{\bf {\color{red}96.5}} & \multicolumn{1}{c|}{83.9} & \multicolumn{1}{c|}{77.8} & \multicolumn{1}{c|}{92.1} &
\multicolumn{1}{c|}{\bf {\color{red}84.8}} &
\multicolumn{1}{c|}{\bf {\color{red}78.0}} & \multicolumn{1}{c|}{\bf {\color{red}93.2}} & {\bf {\color{red}83.6}} & {\bf {\color{red}95.1}} & {\bf {\color{red}78.3}} & {\bf {\color{red}92.8}} & {\bf {\color{red}70.6}} & {\bf {\color{red}89.2}}\\
\hline
\end{tabular}}
\end{center}
\end{table*}

\subsubsection{Parameter analysis of the number of part regions $D$}
In addition, we analyse how the number of part regions $D$ in the part extraction module affects the IR results.
We test the performance with $D=\lbrace 4, 6, 8, 10, 12\rbrace$ of our PGAN on VeRI-776 \xinyu{in Table~\ref{table:partnum-analysis} and on VRIC and VERI-Wild in Table~\ref{table:partnum-analysis-vric-veriwild}.}
\xinyu{The evaluated feature dimension is set to $256$ and $512$.}

As shown in Table~\ref{table:partnum-analysis} \xinyu{and Table~\ref{table:partnum-analysis-vric-veriwild}}, there is a consistent improvement when utilizing the part guidance in our PGAN compared with the baseline model,
which clearly verifies the effectiveness of our PGAN method.
When the part number $D$ is not large, our PGAN can gradually improve the IR performance with the number of part regions increasing.
It shows that the fixed number of part regions is able to narrow down the possible searching regions, which is helpful for focusing on the valid part components.
Besides, our PAM can further improve the effectiveness of the part guidance by applying more concentration on the prominent part regions.
Especially,
when $D$ is changed from $6$ to $8$ on VeRi-776 dataset, the performance can be improved by $3\%$ to $3.6\%$ in mAP
when the feature dimension is $512$ and $3.2\%$ to $3.3\%$ in mAP when the feature dimension is $256$ compared with the baseline model.
When $D=8$, we can obtain the relatively best result.
However, the performance decreases when the part number continually increases.
The reasons are twofold: 1)
many
detected part regions are covered with each other, which provide no further part information for the model learning; 2) more wrongly detected parts are extracted that results in the distraction of the model learning via providing large invalid information.
We believe that if we use a better detector, the performance
will
be further improved.

\xinyu{The similar trend is observed on VRIC dataset, as shown in Table~\ref{table:partnum-analysis-vric-veriwild}.
However, for VERI-Wild dataset, our PGAN is relatively robust to the part number.
The reason is that images in VERI-Wild are high resolution and the part regions are detected more accurately.
A few part regions are satisfactory to distinguish different vehicles.
}
Although there exists an optimal $D$ for a specific dataset, we use $8$ as the default setting for simplification.

\subsubsection{Effectiveness on different baseline}
\begin{figure}[t!]
\centering
\includegraphics[trim =0mm 0mm 0mm 0mm, clip, width=.8\linewidth]{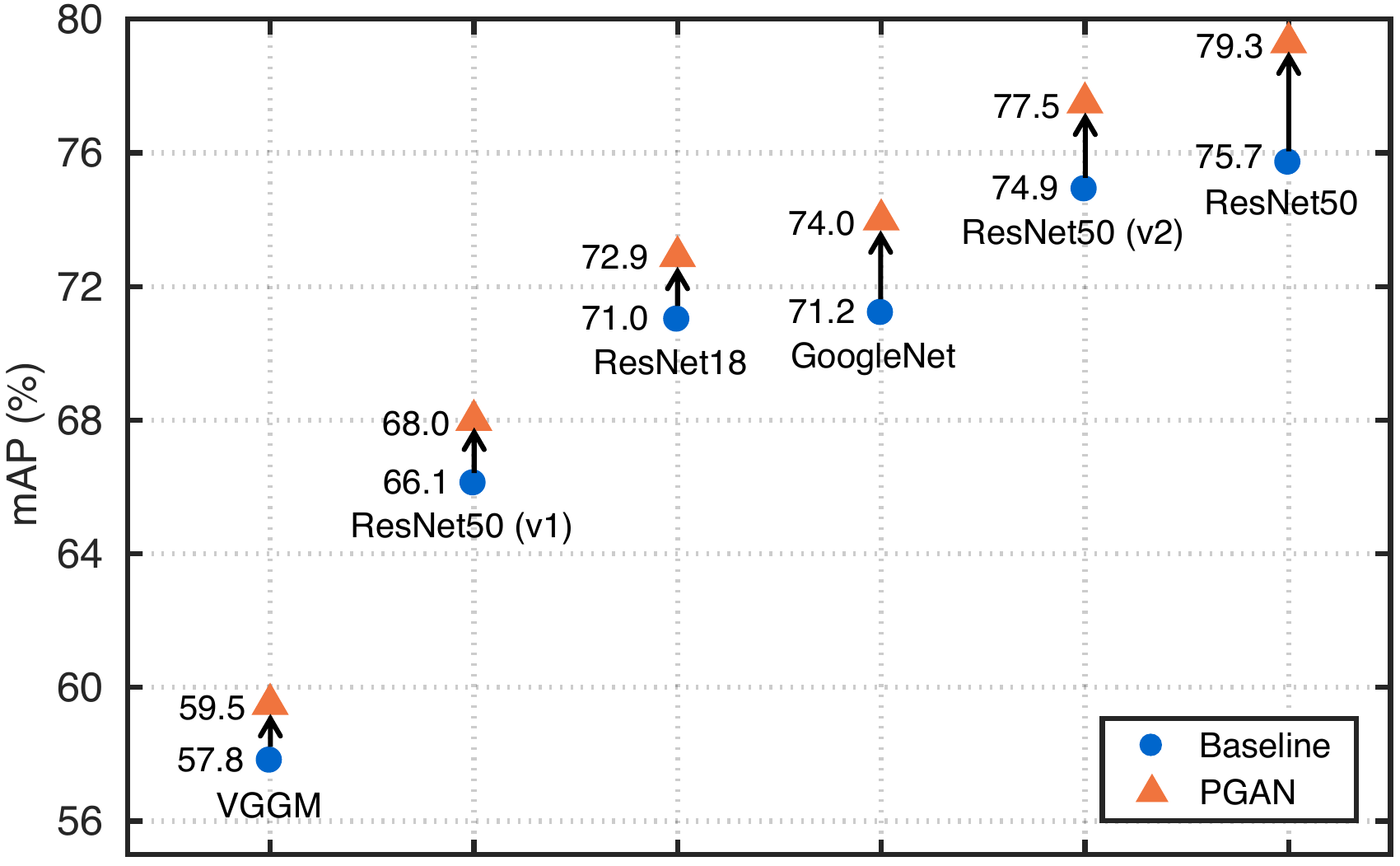}
\caption{Effectiveness on different baseline. v1 and v2 are the variation versions of our used baseline ResNet50 model~\cite{luo2019bag}. v1 denotes the plain ResNet50 (\xinyu{last stride is 2})
in \cite{resnet50} without data augmentation~\cite{randomerasing}. v2 denotes the \xinyu{v1} with \cite{randomerasing}.
Here, we remove the downsampling and use \cite{randomerasing} in ResNet18, GoogleNet and ResNet50 for fair comparison.
}
\label{fig:multibaseline}
\end{figure}

In order to fully verify the effectiveness of our PGAN, we apply our method on various of baseline models.
As shown in Figure~\ref{fig:multibaseline},
\xinyu{we can see that deeper backbone is beneficial for the performance, \eg, ResNet18 baseline achieves 71.0\% mAP while ResNet50 baseline 75.7\%.}
\xinyu{
In particular, we discard the last three FC layers in VGGM~\cite{chatfield2014return} backbone and the last FC layer in GoogleNet~\cite{szegedy2015going} backbone to insert our PGAN into the model.
PGAN gains $2.8\%$ and $1.7\%$ mAP increases when applied on the GoogleNet baseline and VGGM baseline respectively.
This validates the effectiveness of our proposed PGAN, which can be used as a general module for other tasks to some extent.
We can also see that the improvement on VGGM is less than PGAN applied on ResNet50 baseline ($3.6\%$ mAP).
The reason might be that
the size of channels of the output features from the VGGM and GoogleNet backbone is 512 and 1024, which is smaller than that of ResNet50 (\ie, 2048).
Therefore, PAM module has more robust ability of feature representation in ResNet50 than that in VGGM and GoogleNet backbone.
}

Moreover, we can observe that some training methods in~\cite{luo2019bag} are beneficial for the performance increase.
In particular, when we use ResNet50 (v1), the original model with last stride as $2$ and without data augmentation~\cite{randomerasing},
our PGAN obtains about $3\%$ improvement.
Our PGAN is also useful for the version of ResNet50 (v2), \ie, adding \cite{luo2019bag} on ResNet50 (v1).
We can get $2.6\%$ mAP improvement when comparing with the baseline.
Overall, our PGAN can provide consistent improvement whatever the baseline is.

\begin{figure*}[t!]
\centering
\includegraphics[trim =0mm 0mm 0mm 0mm, clip, width=0.875\linewidth]{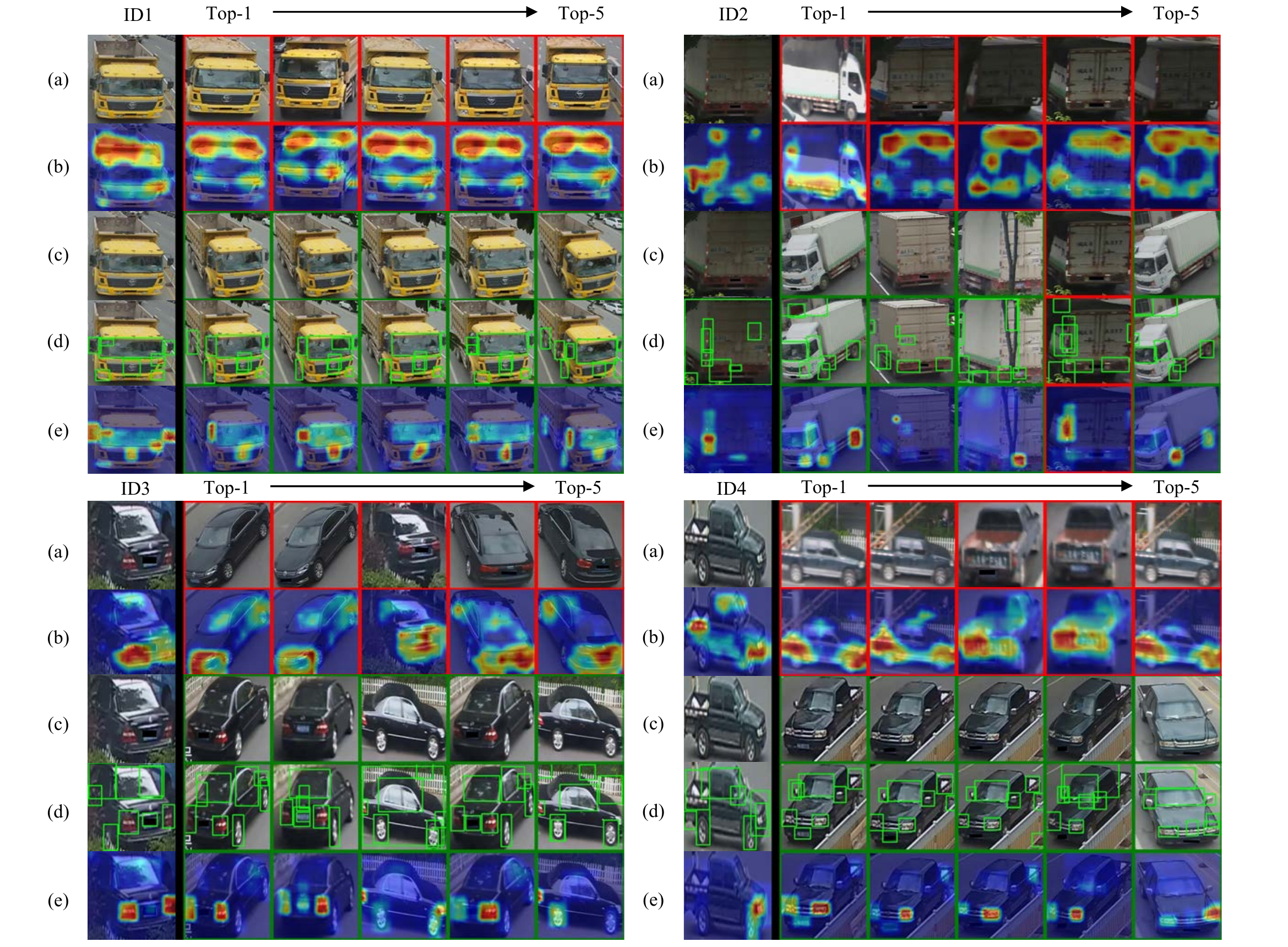}
\caption{Illustration of visualized comparison between traditional grid attention and our PGAN \xinyu{on VeRi-776 dataset}.
For a query image, we draw:
(a) Top-5 retrieval results and (b) the corresponding heatmaps of $\mathbf{F}_p$ from PAM in grid attention;
(c) Top-5 retrieval results, (d) the detected candidate part regions and (e) the corresponding heatmaps of $\mathbf{F}_p$ from PAM in PGAN.
The correct and false matched vehicle images are enclosed in {\color{green!50!black}green} and {\color{red!90!black}red} rectangles respectively.
It shows that our PGAN can put more attention on the most prominent part regions, such as back mirrors, windshield stickers and car brands.
However, the grid attention mainly focuses on some insignificant regions like the car roof, resulting in the attention distracting.
(Best viewed in color)
}
\label{fig:visual-supp}
\end{figure*}

\subsection{Comparison with State-of-the-art Methods}
Finally, we compare our PGAN against other state-of-the-art vehicle IR methods, shown in Table~\ref{table:SOTA}. All reported results of our method are based on $512$-dimension $\widetilde{\mathbf{F}}_{fb}$.

For VeRi-776, we strictly follow the cross-camera-search evaluation protocol as~\cite{liu2016deep}.
From Table~\ref{table:SOTA}, it is clear that our PGAN outperforms all the existing method for a large margin. For instance, the performance of PGAN is better than the state-of-the-art method, \ie, Part-Regular~\cite{he2019part}, for $5\%$ mAP and $2.3\%$ Top-$1$ respectively.
\xinyu{RAM~\cite{liu2018ram} concatenates all the global and local features together as the final representation, which achieves 61.5\% mAP with VGGM backbone.
Similar as RAM, our PGAN can achieve 63.6\% mAP with VGGM when combining $\widetilde{\mathbf{F}}_{fb}$, $\widetilde{\mathbf{F}}_{g}$ and $\widetilde{\mathbf{F}}_{p}$ together.}

For VehicleID, we only report the result of the large test subset on Top-$1$ and Top-$5$.
Our method surpasses almost all the methods
except RNN-HA~\cite{wei2018coarse} at Top-$1$ and VANet~\cite{Chu_2019_ICCV}.
Notice that RNN-HA uses the additional supervision of the vehicle model and the size of input image is $672\times 672$ ($9$ times bigger than ours).
However, as reported in~\cite{wei2018coarse}, the performance of RNN-HA is extremely dropped by a large margin on VeRi-776 when the image size is set to $224\times 224$, which is lower than our PGAN for about $22\%$ in Top-$1$.
In addition, VANet uses a specific viewpoint-based loss function, in which viewpoint labels are generated from a viewpoint model that is trained on manually annotated training samples.
It is specially good for the front and rear view that appears in all vehicles in VehicleID.
For VeRi-776, containing multi-view vehicles,
the mAP of VANet is lower than our PGAN by $13\%$.
Since VANet only reports the result on GoogleNet backbone on VeRi-776,
we also use the same backbone in our PGAN.
From Figure~\ref{fig:multibaseline}, we can see that our PGAN obtains $71.2\%$ mAP using GoogleNet as the backbone model, which is largely higher than VANet ($66.3\%$ mAP).
It means that our PGAN is more beneficial for improving the performance in the multi-view scenario.

For VRIC, one of the largest dataset in vehicle IR, our proposed PGAN achieves satisfactory performance with $78.0\%$ mAP and $93.2\%$ Top-$1$.
Note that we only use single resolution in both training and inference stages, achieving higher performance than MSVR~\cite{vric} using multi-scale feature representations.
\xinyu{We can also observe that although our baseline model has achieved satisfactory results, our PGAN can still improve the performance.
It proves that although suffering from extreme motion blur, low resolution and various complex environment in VRIC dataset, our method can still extract useful and valid information.
Since the detector~\cite{zhao2019structural} is applied without finetune, we believe that with a more accurate detector, our PGAN is able to perform better.
}

VERI-Wild is a newly released large vehicle dataset with more unconstrained variations in resolutions, illuminations, occlusion, and viewpoints, \etc.
There are only a few methods that have reported the results.
Table~\ref{table:SOTA} shows that our proposed PGAN achieves great improvement compared with other methods, \eg, achieving $70.6\%$ mAP at the large test subset.
FDA-Net~\cite{veriwild} uses grid attention module to strengthen the model ability on local subtle differences, which performs worse than our PGAN.
For fairness, we also conduct grid attention instead of PAM in our method and achieve \check{$70.2\%$} mAP, showing that our method is more useful via focusing on subtle differences.
Moreover, we apply VGGM as the backbone model for VeRi-776, as shown in Figure~\ref{fig:multibaseline}.
Our PGAN gets $59.5\%$ mAP that is better than FDA-Net ($55.5\%$).

\xinyu{We also report the result of the baseline model on all dataset. Note that, we also set the feature dimension to $512$ in baseline model for fairness since the feature dimension of our fusion feature $\widetilde{\mathbf{F}}_{fb}$ is set to $512$.
Experiments show that our method achieves higher results than the baseline model in all datasets.}

\begin{figure}[t!]
\centering
\includegraphics[trim =0mm 0mm 0mm 0mm, clip, width=0.985\linewidth]{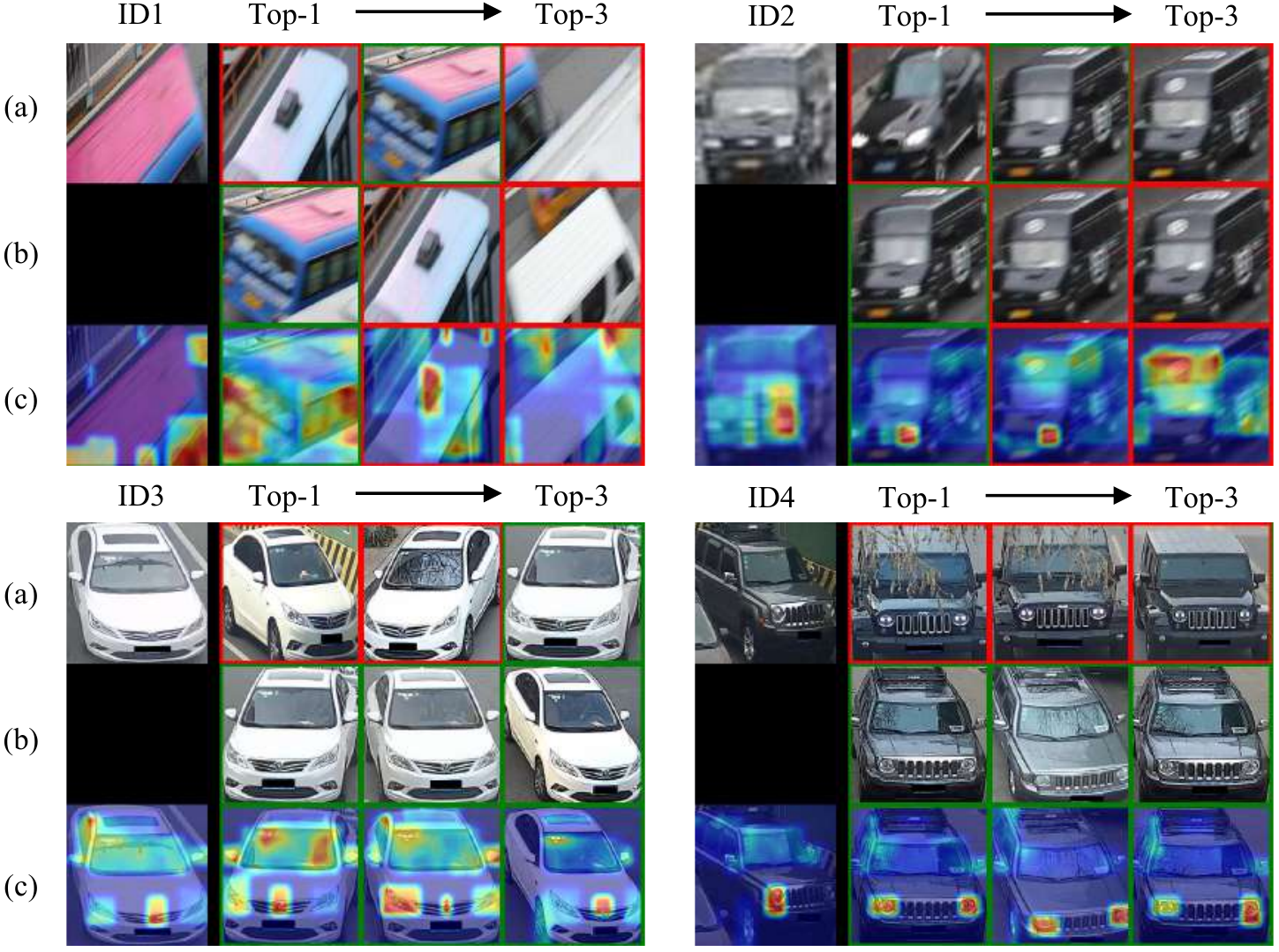}
\caption{Visualization of Top-3 retrieval images of baseline and our PGAN.
ID1 and ID2 are from VRIC dataset, while ID3 and ID4 are from VERI-Wild dataset.
For a query image, we draw:
(a) Top-3 retrieval results from baseline;
(b) Top-3 retrieval results and (c) the corresponding heatmaps of $\mathbf{F}_p$ from PAM in PGAN.
The correct and false matched vehicle images are enclosed in {\color{green!50!black}green} and {\color{red!90!black}red} rectangles respectively.
}
\label{fig:visual-vric-veriwild}
\end{figure}

\subsection{Visualization}
\xinyu{In this section, we visualize some retrieval results of the baseline, grid attention and our part-guided attention method (PGAN), repectively. As shown in Figure~\ref{fig:visual-supp}, we illustrate four different query vehicle images and their corresponding Top-$5$ most similar images as well as the heatmaps of $\bF_p$ from the gallery set on VeRi-776 dataset.
Meanwhile, we illustrate the Top-$3$ retrieval results on VRIC and VERI-Wild datasets in Figure~\ref{fig:visual-vric-veriwild} to show the effectiveness of our PGAN.}
In detail, the main advantages of our PGAN can be summarized as follows:

\subsubsection{Insensitive to various situations} Our
PGAN can extract more robust feature representation so as to  significantly improve the IR performance. As shown in the ID2 and ID3 in Figure~\ref{fig:visual-supp}, given a rear vehicle image, we can not only find the easy vehicles from the rear views, but also get the side-view vehicle images that are difficult to recognize even by humans.
In contrast, the grid attention can only focus on the images from the nearly same views.
Moreover, our PGAN is also able to deal with various situations.
\xinyu{As shown in Figure~\ref{fig:visual-vric-veriwild}, although images in VRIC and VERI-Wild datasets suffer from blur, illumination and occlusion, our PGAN can still find the correct vehicles according to the prominent part regions.}
It means that our method is more robust to learn discriminative features that is not sensitive to multiple variants from the environment.
\subsubsection{The effectiveness of the part extraction module as the bottom-up attention} The detected part regions play an important role in feature representation. As illustrated in the ID3, it is clear that the wrongly retrieved images from the grid attention method are different from the query image from the car lights.
However, a lot of regions representing the body and the bottom of the car are concentrated, which are not the obvious differences between two vehicles.
Nevertheless, with the guidance of the detected part regions, our PGAN can only focus on these candidate regions that is beneficial for focusing on useful regions as well as alleviating the bad effect from the other regions.
In other words, the part extraction module helps the network learning by narrowing down the searching ranges.

\begin{figure}[t!]
  \begin{minipage}[t!]{0.23\textwidth}
    \footnotesize
    \centering
    \tabcaption{The average running time (ms) per frame on VeRi-776.}
    \setlength{\tabcolsep}{0.1mm}{
    \begin{tabular}{l|c|c|c}
    \hline
    \multirow{2}{*}{Method}        & \multicolumn{3}{c}{Module} \\ \cline{2-4}
                                   & Detector  & IR  & Total  \\ \hline \hline
    \multicolumn{1}{c|}{Baseline} & -         & 11.8   & 11.8   \\ \hline
    \multicolumn{1}{c|}{PGAN}     & 39.2      & 12.3   & 51.5   \\
    \hline
    \end{tabular}}
    \label{table:runningtime}
  \end{minipage}
  \begin{minipage}[t!]{0.245\textwidth}
    \centering
    \includegraphics[width=1.0\textwidth]{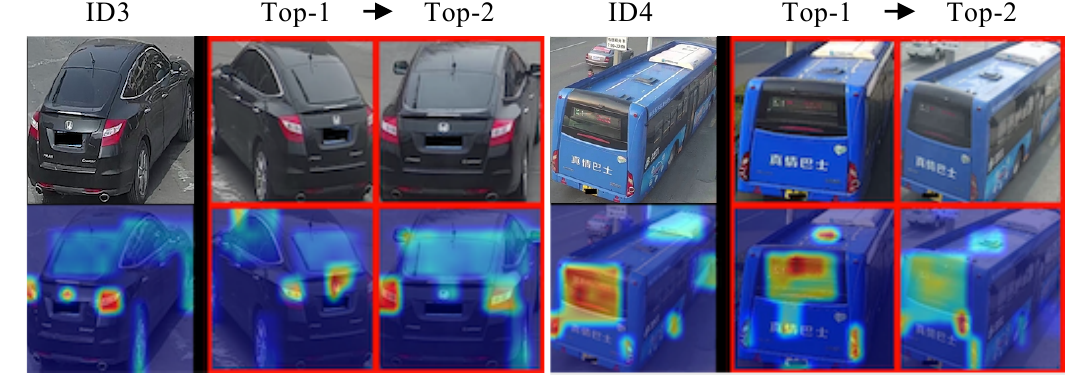}
    \setlength{\belowcaptionskip}{-0.0cm}
    \setlength{\abovecaptionskip}{-0.1cm}
    \caption{Illustrations of some failed samples on the  VRIC dataset.}
    \label{fig:liminitation}
  \end{minipage}
\end{figure}

\subsubsection{The effectiveness of the part attention module as the top-down attention} Our PGAN is useful for selecting the most prominent part regions and lighten the influence of invalid and useless regions. As described in the main paper, we propose a part attention module (PAM) that is responsible for learning a soft attention weight for each part.
Therefore, the important part regions are underlined by a high-attention value, while the impact of other insignificant parts is relatively suppressed.
From the feature maps, we can clearly observe that our PGAN could focus on the most prominent part regions, such as the car lights in ID3, back mirrors in ID1.
As shown in ID4, although there are few valid part regions that are extracted, our PGAN can still find the key information to recognize the vehicles, such as the wheel and the cat lights.
On the contrary, the grid attention is largely influenced by some invalid regions that are extremely similar in different vehicles, such as the bottom of the vehicle body.
\xinyu{\subsubsection{The influence of the overlapped part regions}
When the area of the overlapped region is large, the attention weights of both two regions from our PAM will be tended to be large consistently if this region is prominent, and vice versa, \eg, the side wind-shield glass of ID3 in Figure~\ref{fig:visual-supp}.
For another situation that the area of the overlapped region is small,
our PAM can provide the overall evaluation for every part region.
For example, as the illustrated image in Figure 3, the attention weight of the annul services sign is large due to its unique, while the weight of the wind-shield glass is small because it has relatively less informative information.
}
\xinyutwo{\subsubsection{The limitation of our PGAN}
From Figure~\ref{fig:liminitation}, it shows that our PGAN fails to distinguish: i) the extreme similar vehicles that share the same appearance; ii) the public vehicles without unique features.
It is reasonable since our PGAN depends on the discriminative information.
If vehicle plates are available, our PGAN can achieve higher performance.
}

\begin{figure}[t!]
\centering
\includegraphics[trim =0mm 0mm 0mm 0mm, clip, width=0.98\linewidth]{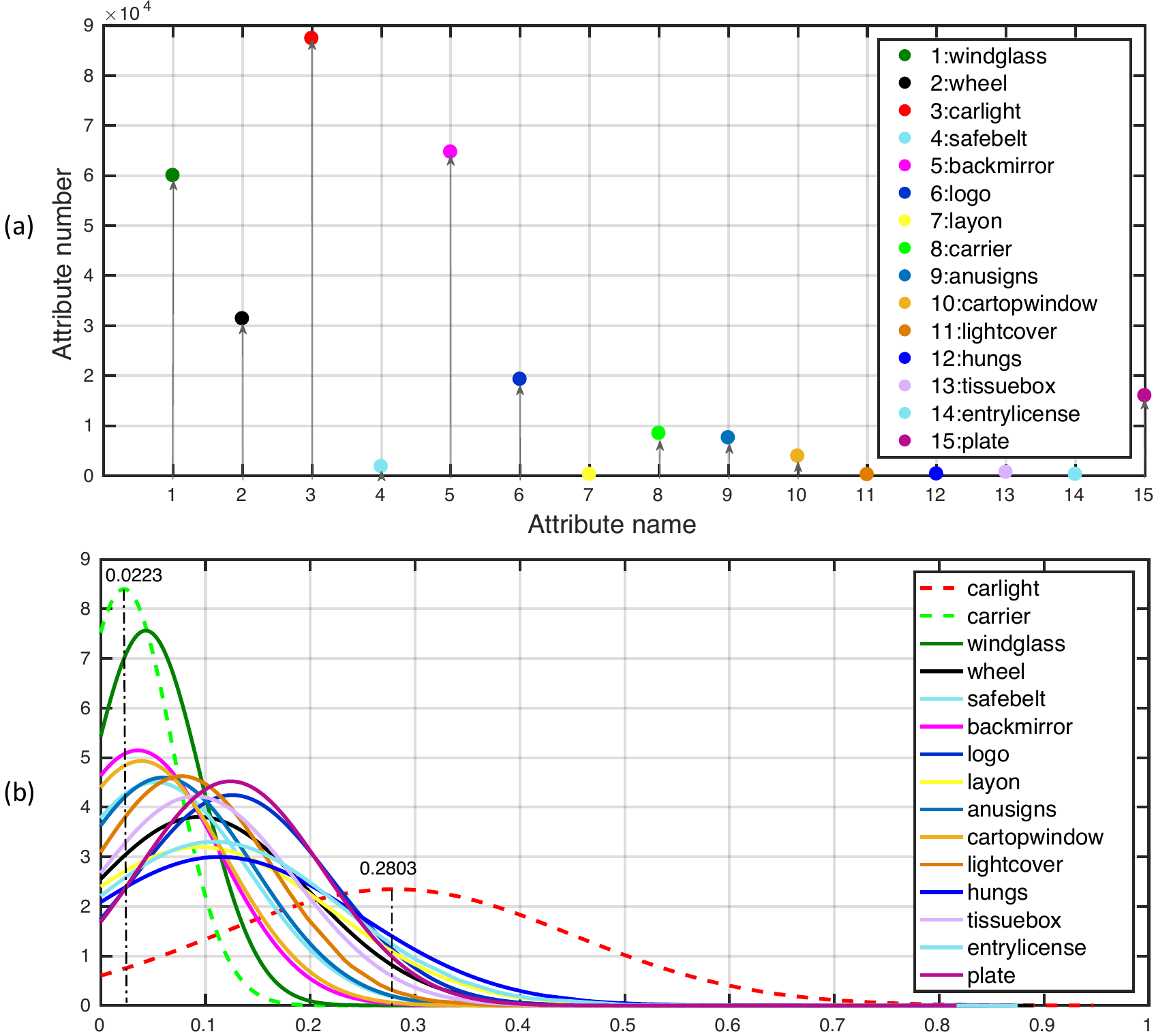}
\caption{\xinyu{Statistical analysis of the effectiveness of each attribute in Table~\ref{table:attributes} on VeRi-776 training dataset.
Attribute names are denoted as the abbreviations in Table~\ref{table:attributes}.
(a) Number statistic for each vehicle attribute.
(b) The probability density function of attention weights from PAM for each attribute.
The most prominent attribute is shown in red dashed line, while the least informative attribute is shown in green dashed line.
Best view in color.}
}
\label{fig:attribute-analysis}
\end{figure}

\xinyu{
\subsection{Discussion}
As shown in Table~\ref{table:attributes}, 16 attributes are included in our part extraction module.
Although the attribute information is ignored when selecting the top-$D$ part regions based on the confidence scores, we can still analyze which part regions are prominent.
We extract the attention weights from the PAM on the VeRi-776 training dataset.
Note that we set $D=8$ in this section and newer sign attribute is not appeared in VeRi-776 training dataset.
From Figure~\ref{fig:attribute-analysis}(a), we can observe that carlight is the most frequently selected part.
It makes sense that carlight appears in almost all vehicles whatever the vehicle view is.
Moreover, windglass, backmirror and wheel also appear frequently, while layon (lay ornament), entry license, hungs and tissue appear rarely.
Refer to Figure~\ref{fig:attribute-analysis}(b), it is clear that carlight plays the most important role in distinguishing different vehicles.
It is interesting to see that some subtle part regions still have useful information, such as logo, hungs, entrylicense and annusigns, although these attributes appear less than windglass and backmirror that include
nearly no
information.
The analyses show that our PGAN is effective to attend the meaningful and useful information for identifying vehicles in an interpretatable way.
}

\xinyutwo{
Furthermore, we also report the running time. All experiments are conducted on a GeForce GTX 1080 Ti machine.
Table~\ref{table:runningtime} shows that the IR module in PGAN can achieve the comparable speed with the baseline despite the additional PAM module and feature aggregation module.
Although the SSD detector~\cite{zhao2019structural} is time-consuming, our PGAN is still practical in the real world.
}

\section{Conclusion}
In this work, we have presented a novel Part-Guided Attention Network (PGAN) for vehicle instance retrieval (IR).
First, we extract part regions of each vehicle image from an object detection model.
These part regions provide a range of candidate searching area for the network learning,
which is regarded as a bottom-up attention process.
Then we use the proposed part attention module (PAM) to discover the prominent part regions by learning a soft attention weight for each candidate part, which is a top-down attention process.
In this way, the most discriminative parts are highlighted with high-attention weights, while the opposite effects of invalid or useless parts are suppressed with relatively low weights.
Furthermore, with the joint optimization
of
the holistic feature and the part feature, the IR performance can be further improved.
Extensive experiments show the effectiveness of our method.
The proposed PGAN outperforms other state-of-the-art methods by a large margin.
We plan to extend the proposed method to the multi-task learning, \ie, object detection and tracking, for simultaneously improving the performance of these two tasks.

\section*{Acknowledgment}
X. Zhang, R. Zhang, and M. You
were  in part
supported by the Shanghai Natural Science Foundation under grant no.\  18ZR1442600, the National Natural Science Foundation of China under grant no.\  62073244 and the Shanghai Innovation Action Plan under grant no.\  20511100500.
\ifCLASSOPTIONcaptionsoff
  \newpage
\fi

{
\bibliographystyle{IEEEtran}
\bibliography{egbib}
}
\end{document}